\documentclass{article} 
\usepackage{iclr2027_conference,times}

\usepackage{amsmath,amsfonts,bm}

\def\eqref#1{equation~\ref{#1}}

\def\1{\bm{1}}

\DeclareMathAlphabet{\mathsfit}{\encodingdefault}{\sfdefault}{m}{sl}
\SetMathAlphabet{\mathsfit}{bold}{\encodingdefault}{\sfdefault}{bx}{n}

\usepackage{fvextra,float}
\usepackage{hyperref}
\usepackage{url}
\usepackage{booktabs}
\usepackage{graphicx}
\usepackage[table]{xcolor}
\usepackage{comment}
\usepackage{wrapfig}
\usepackage{mdframed}
\usepackage{tcolorbox}

\newtcolorbox{insbox}{colback=yellow!8, colframe=yellow!55!black,
  boxrule=0.6pt, left=6pt, right=6pt, top=4pt, bottom=4pt, arc=2pt}

\title{Fewer Tokens, More Self-Teaching: On-Policy Self-Distillation for Extreme Visual Token Reduction}

\author{Junxian Li$^{1}$\thanks{These authors contributed equally.} \
Ruixuan Yang$^{2*}$ \
Tianao Zhang$^{1}$ \
Tiange Xu$^{3}$  \ \textbf{Weisheng Dong}$^{4}$ \ \textbf{Yulun Zhang}$^{1}$\thanks{Corresponding author.} \\
$^{1}$Shanghai Jiao Tong University, $^{2}$Xi'an Jiaotong University, \\ 
$^{3}$University of Cambridge, $^{4}$Xidian University
\\
\texttt{
lijunxian0531@sjtu.edu.cn, yulun100@gmail.com}
}

\iclrfinalcopy 
\begin{document}

\maketitle
\fancyhead[L]{Preprint} 

\begin{abstract}
Visual token reduction is an effective way to accelerate multimodal large language models (MLLMs), but performance deteriorates rapidly under extremely low token budgets. Existing work has explored both visual-token selection and training-based adaptation to reduced visual inputs. We take a step further by asking how a heavily compressed MLLM should learn from the states induced by its own generations. This setting naturally calls for on-policy self-distillation: a heavily compressed model is supervised on the states induced by its own generations, while its full-token counterpart serves as an information-rich teacher. Based on this insight, we propose LT-OPD, a training framework for extreme visual-token reduction. The student rolls out responses with only a small fraction of visual tokens, and a frozen full-token copy of the same MLLM provides distributional supervision along these student-generated trajectories. To stabilize on-policy learning when visual evidence is severely limited, we further introduce a budget-level curriculum that progressively decreases the token budget during training. Across nine benchmarks on Qwen3.5-4B, LT-OPD raises average retained performance under 5\% visual-token retention from 68.6\% to 82.3\%, outperforming training-free, training-based, and reinforcement-learning baselines at the same budget. The gains transfer consistently to Qwen3.5-9B, GLM-4.6V-9B, and LLaVA-OV-1.5-4B. LT-OPD also reduces KV-cache usage by 85.2\% and prefill FLOPs by 85.4\% without additional inference overhead, demonstrating that on-policy learning can substantially recover capabilities lost to extreme visual-token reduction. Code is at \url{https://github.com/Yrxxxxxxxx1007/LT-OPD}.
\end{abstract}

\vspace{-3mm}
\section{Introduction}
\vspace{-3mm}

Multimodal large language models (MLLMs)~\citep{liu2023visual,hurst2024gpt,bai2025qwen3,team2025kimi,wang2025internvl3,an2025llava,hong2025glm,li2025chemvlm} have attracted research interests for years. While exploring the frontiers of MLLMs in many aspects like multimodal understanding and planning, visual question answering and visual agentic planning, the efficiency of such models also matters. When deployed on edge devices or used in resource-constrained scenarios, deploying such MLLMs with billions of parameters remains challenging.

\begin{figure*}[t]
    \centering

    \begin{minipage}[t]{0.332\linewidth}
        \centering
        \includegraphics[width=\linewidth]{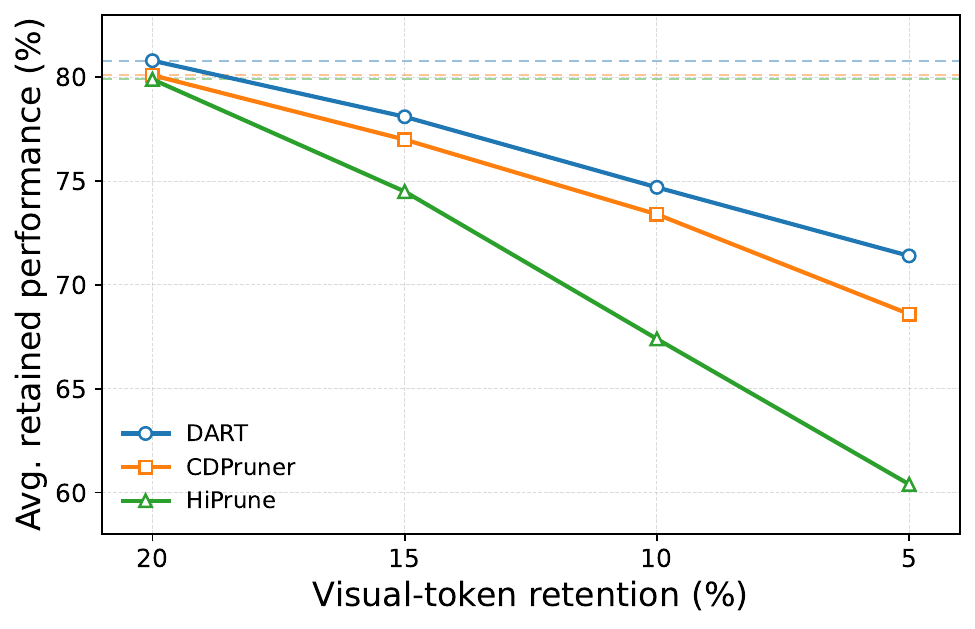}
    \end{minipage}
    \hfill
    \begin{minipage}[t]{0.31\linewidth}
        \centering
        \includegraphics[width=\linewidth]{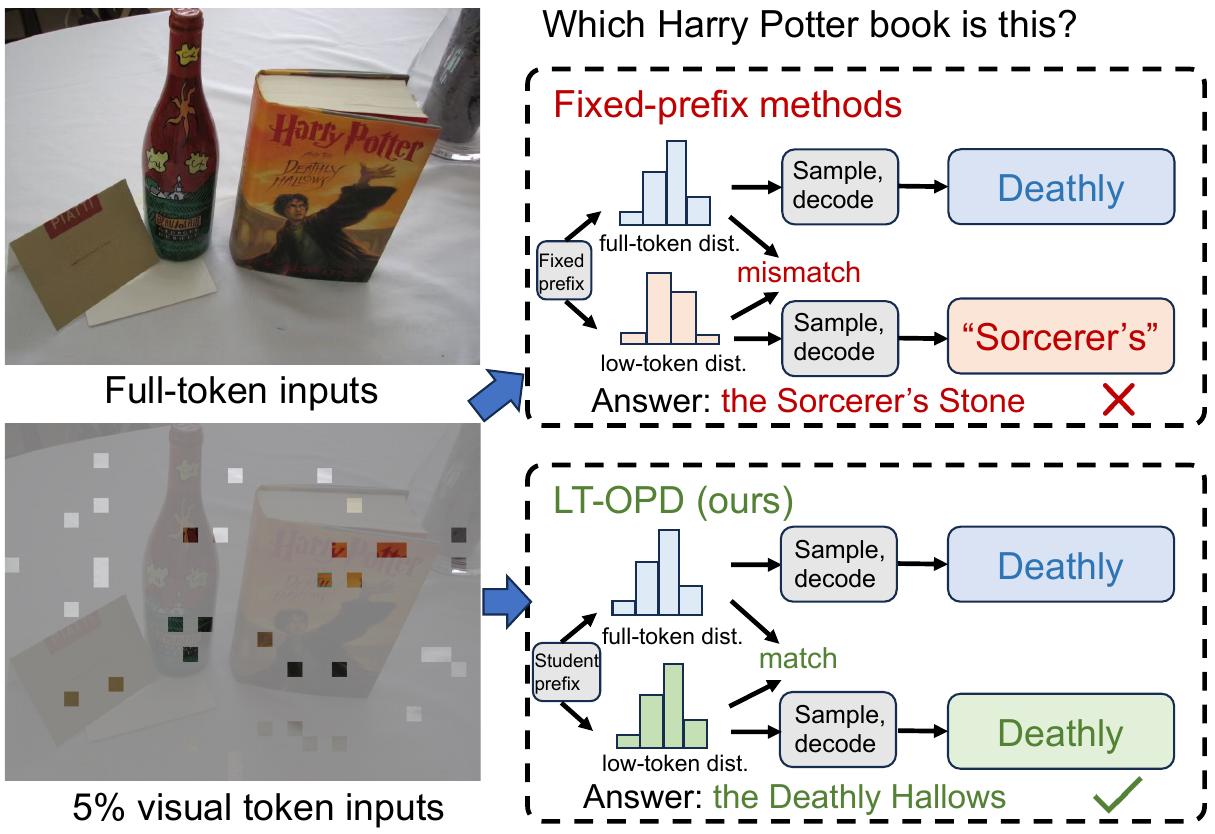}
    \end{minipage}
    \hfill
    \begin{minipage}[t]{0.31\linewidth}
        \centering
        \includegraphics[width=\linewidth]{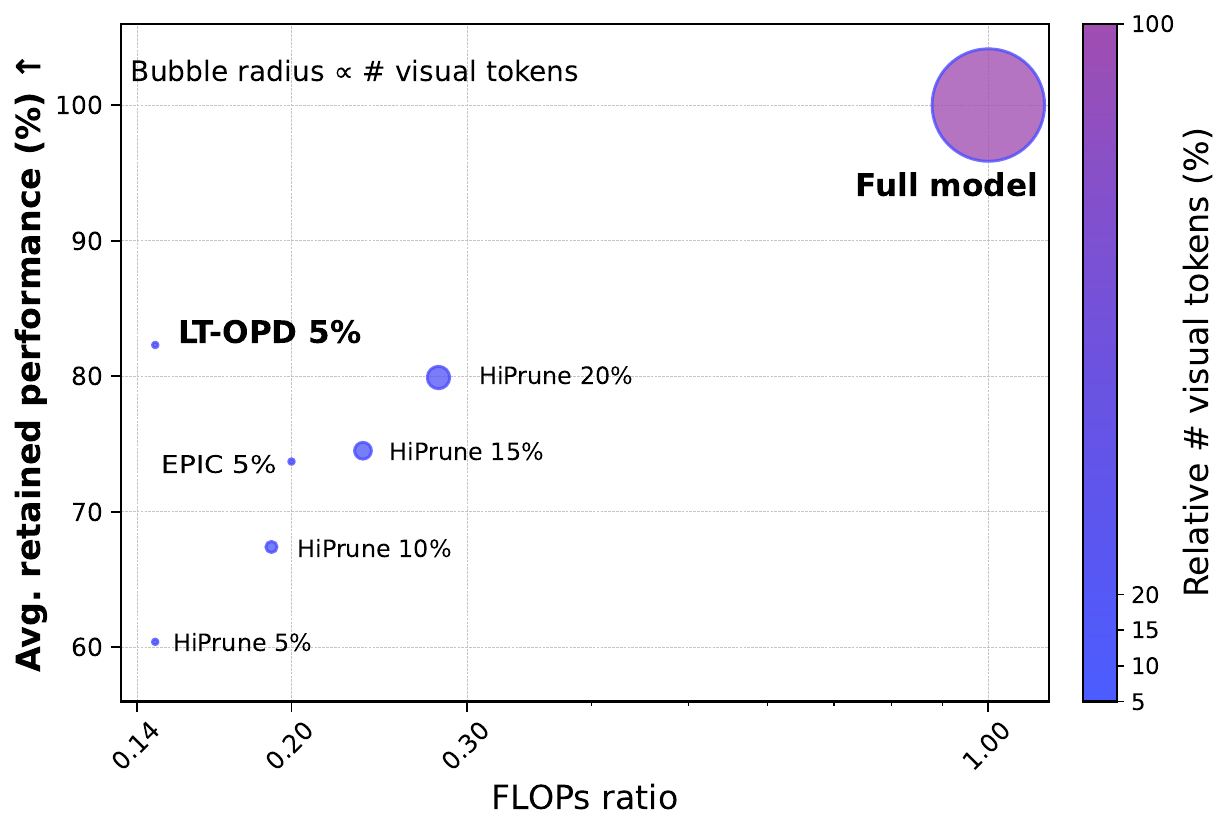}
    \end{minipage}
    \vspace{-3mm}
    \caption{Motivation and performances of LT-OPD. Left: different visual token reduction methods all fail to perform well under an extremely low budget. Middle: Under this scenario, predictive distribution mismatch caused by fixed-prefix methods can lead to the wrong answer. LT-OPD, however, leverages on-policy learning to mitigate the problem. ``dist.'' means distribution and ``student prefix'' means prefix generated by students. Right: Compared with HiPrune (5\%, 10\%, 15\%, 20\%), and EPIC (5\%), our LT-OPD reaches the best trade-off.}
    \vspace{-6mm}
    \label{fig:motivation}
\end{figure*}

To address this challenge, substantial efforts have been made. Visual token reduction~\citep{chen2024image,alvar2025divprune,zhang2025beyond,wen2025stop,li2026g}, which reduces the visual tokens from the visual encoder, acts as a training-free and usually plug-and-play strategy in increasing MLLMs' efficiency. However, another problem occurs on applying these training-free techniques. As Figure~\ref{fig:motivation} left suggests, an extremely low visual token budget (ratio of preserved tokens) may be set, like 5\%, due to serious resource limits. Under this scenario, focusing on \textbf{\textit{which visual tokens should be retained}} cannot achieve acceptable performances of MLLMs. 

We argue that, to achieve better performances under extremely low budgets, exploring \textbf{\textit{how to adapt a model to operate effectively}} with severely compressed visual inputs also matters. And this needs special adaptive training. Several previous works~\citep{hu2024matryoshka,zhang2025llava,wen2025efficient} support this idea and make efforts on developing training techniques like finetuning with pre-fusion and consistency training. While obvious improvements are seen, a remaining limitation prevents them from achieving further gains. We observe that, low token budgets change not only the amount of visual evidence available, but also the generation trajectories encountered at inference time. 
As illustrated in Figure~\ref{fig:motivation} middle, with severely limited visual inputs, the predictive distribution of the compressed model can deviate substantially from that of its full-token counterpart. Training directly
on fixed response prefixes does not account for the states from the compressed model's own predictions. This mismatch becomes particularly relevant once generated tokens are fed back as subsequent context. This observation naturally motivates us to adapt the compressed model on the trajectories it actually generates, rather than only learning from fixed training prefixes.

Based on this insight, we introduce \textbf{L}ow-\textbf{T}oken \textbf{O}n-\textbf{P}olicy \textbf{D}istillation (\textbf{LT-OPD}), which contains two complementary components. First, we perform a specific on-policy self-distillation algorithm: the low-token student rolls out its own responses. Then a frozen full-token counterpart provides distributional supervision along the prefixes generated by students. In this way, the student learns to recover the behavior of the information-rich teacher exactly at the states it is likely to encounter during inference, as Figure~\ref{fig:motivation} middle shows. Second, directly conducting on-policy learning under an extremely low token budget can be difficult, since the heavily compressed student may struggle to generate useful trajectories at the beginning of training. We therefore introduce a budget-level curriculum. It starts training with a relatively larger budget and progressively reduces it to the target budget. This gradually shifts the student from an easier visual condition to the desired highly compressed, harder condition. This leads to more stable and effective adaptation.

Extensive experiments demonstrate the effectiveness and generality of LT-OPD. Across nine benchmarks, LT-OPD retains 82.3\% of the full performance on Qwen3.5-4B while preserving only 5\% of visual tokens, substantially improving over the corresponding untrained compressed model (68.6\%) and the strongest baseline (77.6\%) under the same budget. The improvement consistently transfers across model scales and architectures. 
Importantly, as Figure~\ref{fig:motivation} right depicts, these gains come without decreasing inference efficiency: the same 0.146$\times$ FLOPs ratio as HiPrune~\citep{liu2026hiprune} while improving retained performance by 21.9 percentage points, and 8.6 points higher with 27.0\% fewer FLOPs than EPIC~\citep{wen2025efficient}. In sum, our contributions can be summarized as:

$\bullet$ We revisit extreme visual-token reduction from a training perspective. 
    We highlight that severe visual compression not only removes visual evidence, but also changes the generation states encountered by the compressed model, motivating adaptation directly on its own generated trajectories.

$\bullet$ We propose \textbf{LT-OPD}, an on-policy self-distillation framework under this scenario. LT-OPD lets the compressed student learn from a 
    full-token self-teacher on states induced by student, and further introduces a 
   budget-level curriculum that progressively reduces the budget to stabilize learning.

$\bullet$ We conduct extensive evaluations across nine benchmarks, multiple model scales, and three MLLM families. With only \textbf{5\%} visual tokens, LT-OPD outperforms a wide range of baselines.

\vspace{-3mm}
\section{Related Works}
\vspace{-3mm}
\textbf{Token Reduction for Efficient MLLMs.} 
\begin{table}[t]
\centering
\caption{
Comparison of representative training-based visual-token reduction methods.
``Fixed prefix'' means that generation states are determined by static training
responses rather than rollouts of the current compressed model.
}
\label{tab:training_paradigm}
\small
\setlength{\tabcolsep}{3.5pt}
\renewcommand{\arraystretch}{1.12}
\begin{tabular}{l p{7.0cm} l c}
\toprule
\textbf{Method}
& \textbf{Training Strategy}
& \textbf{Prefix / State Source}
& \textbf{On-policy} \\
\midrule
MQT-LLaVA
& Randomly samples the number of query tokens and optimizes the language-modeling objective
& Fixed training
& No \\

LLaVA-Mini
& Trains compression and modality pre-fusion modules through end-to-end instruction tuning
& Fixed training
& No \\

EPIC
& Progressive token/layer consistency distillation with SFT and KL supervision
& Fixed training
& No \\

LearnPruner
& Freezes the base VLM and trains the pruning module with next-token prediction and pruning losses
& Fixed training
& No \\

\rowcolor{orange!10}
\textbf{LT-OPD}
& \textbf{Rolls out a low-token student and distills a frozen full-token self-teacher with a budget curriculum}
& \textbf{Student-generated}
& \textbf{Yes} \\
\bottomrule
\end{tabular}
\vspace{-5mm}
\end{table}
Token reduction is a widely used strategy in enhancing MLLM efficiency. It reduces redundant visual tokens, and therefore mainly reduces prefill time of MLLM inference. Two categories of token reduction methods are developed: training-free methods and training-based ones. For training-free methods, FastV~\citep{chen2024image} first utilizes attention scores to determine redundant tokens. Methods like DivPrune~\citep{alvar2025divprune}, DART~\citep{wen2025stop}, CDPruner~\citep{zhang2025beyond}, HiPrune~\citep{liu2026hiprune} and G$^2$TR~\cite{li2026g} further integrate diversity, duplication and so on to judge redundant tokens during reduction. Besides token selection, training-based methods concentrate on teaching the MLLMs with fewer visual tokens how to generate responses similar to those with full visual information. MQT-LlaVA~\citep{hu2024matryoshka}, LlaVA-Mini~\citep{zhang2025llava}, EPIC~\citep{wen2025efficient} are representative methods. LearnPruner~\citep{takezoe2026learnpruner} learns a
task-aware pruning module with next-token prediction supervision. These methods, while achieving notable success, mainly focus on choosing which visual tokens, or training with fixed prefix, as Table~\ref{tab:training_paradigm} depicts. This may stop them from further improvement under our scenario.

\textbf{On-policy Distillation.}
On-policy distillation (OPD) lets the student model optimize on its self-generated trajectories, with supervision from teacher model by token-level KL divergence or other guidances\citep{agarwal2024policy,xu2026deepseek,hou2026uni,li2026rethinking,xu2025speculative}. By this method, students can learn better from training distribution than off-policy distillation~\citep{pmlr-v267-chu25c}. Without online exploration, inference-time prefixes may differ from those seen during training, leading to accumulated errors. Recent works such as OPSD~\citep{zhao2026self} utilizes on-policy distillation on LLMs, with using ground-truth as rewards for optimization. Vision-OPD~\citep{yuan2026vision} further provides self-teacher with fine-grained visual information to facilitate with on-policy distillation on MLLMs. In sum, previous OPD methods aim at enhancing the policy's performance through stronger teachers or teacher with more information. Based on this, we explore whether OPD can help train a well-performing student with fewer visual tokens preserved. 

\vspace{-2mm}
\section{Methodology}
\label{sec:method}
\vspace{-2mm}
LT-OPD is built on a simple framing: a full-token model has access to
richer visual evidence for correcting its compressed counterpart, while the
compressed model defines the generation states that will actually be encountered
at inference time. We therefore let a low-token student generate its own
responses and query a frozen full-token copy of the same MLLM for dense
distributional supervision along these student-induced trajectories.  
Figure~\ref{fig:framework} gives an overview.

\subsection{Problem formulation}
\begin{figure*}[t]
    \centering
    \includegraphics[width=\linewidth]{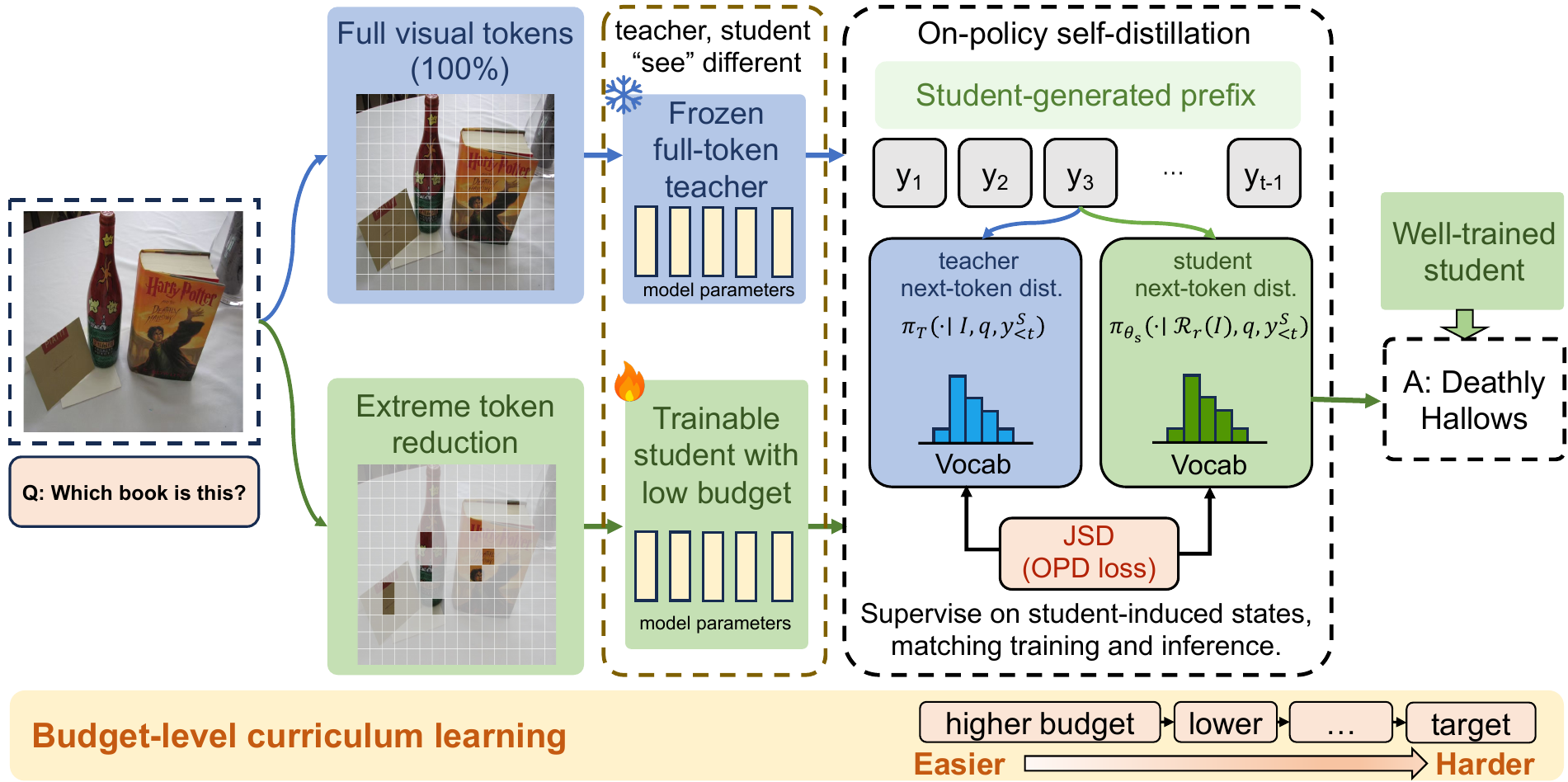}
    \vspace{-4mm}
    \caption{Framework of LT-OPD, consisting of two parts. First, an on-policy self-distillation algorithm is proposed to train a low-token student, with a full-token frozen self-teacher. ``dist.'' means distribution. Second, we design a budget-level curriculum learning process for stable training.}
    \label{fig:framework}
    \vspace{-4mm}
\end{figure*}

\vspace{-2mm}
\label{sec:formulation}
As Figure~\ref{fig:framework} depicts, given an image-question pair $x=(I,q)$, let $\pi_{\theta_0}$ denote the pretrained MLLM and $N$ be the number of visual tokens primarily produced by the visual encoder for $I$.
We instantiate a frozen full-token teacher $\pi_T$ from $\theta_0$ and a student $\pi_S$ with trainable parameters $\theta$, initialized from the same parameters. For a keep ratio $r$, a visual-token reduction operator
$\mathcal{R}_r$ retains
\begin{equation}
N_{stu} = \operatorname{round}(r*N)
\end{equation}
visual tokens for the student, whereas the teacher receives all $N$ tokens.
Our optimization goal is to adapt the visual tokens of $\pi_S$ to a target extreme budget
$r=\beta$ (e.g., $5\%$) while recovering as much of the full-token behavior as possible.

\subsection{Our Method: On-policy Self-Distillation across Budgets}
\label{sec:core_methods}

Current distillation methods like~\citep{wen2025efficient} evaluates the teacher and student on response
prefixes provided by a fixed training distribution. Under extreme visual
compression, however, the student may quickly deviate from these prefixes,
and subsequent predictions are conditioned on states induced by its own
previous outputs. LT-OPD instead performs distillation directly on this
student-induced state distribution.

Let $S$ be the marker for the ``student'' and $T$ for the ``teacher''. At training update $s$, the low-token student, with parameters $\theta_s$ at update $s$, first samples a response
from its current policy,
\begin{equation}
{y}^S=({y}^S_1,\ldots,y^S_T)
\sim
\pi_{\theta_s}(\cdot \mid \mathcal{R}_r(I),q).
\end{equation}
The sampled sequence is treated as a stop-gradient trajectory, i.e., we do not
differentiate through the discrete sampling process. At each decoding step $t$,
the teacher and student are evaluated on the same student-generated prefix
${y}^S_{<t}$, but with different amounts of visual evidence:
\begin{align}
p^S_{t,\theta}
&=
\pi_{\theta_s}
\bigl(\cdot \mid \mathcal{R}_r(I),q,{y}^S_{<t}\bigr),\\
p^T_t
&=
\pi_T
\bigl(\cdot \mid I,q,{y}^S_{<t}\bigr).
\end{align}
We minimize the discrepancy between these two next-token distributions $\pi_{\theta_s}$ and $\pi_T$.
Following prior works~\citep{hou2026uni,yuan2026vision}, we use the
Jensen-Shannon divergence (JSD)~\citep{menendez1997jensen}:
\begin{align}
m_{t,\theta}
&=
\frac{1}{2}(p^T_t+p^S_{t,\theta}),\\
\ell_t^{\mathrm{JSD}}(\theta)
&=
\frac{1}{2}D_{\mathrm{KL}}(p^T_t\Vert m_{t,\theta})
+
\frac{1}{2}D_{\mathrm{KL}}(p^S_{t,\theta}\Vert m_{t,\theta}).
\end{align}

Accordingly, at update $s$ we optimize the following on-policy distillation
surrogate:
\begin{equation}
\widetilde{\mathcal{L}}_{\mathrm{OPD}}^{(s)}(\theta;r)
=
\mathbb{E}_{(I,q)\sim\mathcal{D}}
\mathbb{E}_{{y}^S\sim
\operatorname{sg}\!\left[
\pi_{\theta_s}(\cdot\mid\mathcal{R}_r(I),q)
\right]}
\left[
\frac{1}{|y^S|}
\sum_{t=1}^{|y^S|}
\ell_t^{\mathrm{JSD}}(\theta)
\right],
\label{eq:opd}
\end{equation}
where $\operatorname{sg}[\cdot]$ indicates that the rollout distribution is
treated as fixed during differentiation. Gradients are taken with respect to
the student distributions and evaluated at $\theta=\theta_s$; no
score-function gradient is propagated through the sampled tokens. The teacher
remains frozen throughout training.

Importantly, the full-token teacher does not generate a separate trajectory.
Instead, it provides distributional supervision on prefixes generated by the
current compressed student. Thus, ``on-policy'' refers to the source of the
training states: supervision is performed on states induced by the current
student policy, rather than on fixed response prefixes. After training, only
the compressed student is retained for inference. For implementation efficiency, we use the
top-K vocabulary support described in Appendix~\ref{app:implementation};
objective ablations are provided in Appendix~\ref{app:objectives}.

\subsection{Budget-level Curriculum Learning} 
\begin{wrapfigure}{r}{0.46\columnwidth}
    \centering
    \setlength{\columnsep}{4pt}
    \includegraphics[width=\linewidth]{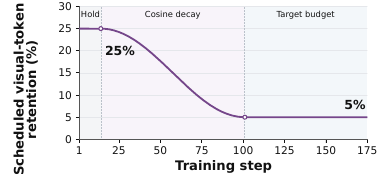}
    \vspace{-6mm}
    \caption{Visual token keep ratio changes.}
    \label{fig:token}
    \vspace{-12pt}
\end{wrapfigure}

On-policy supervision is useful only at states that the student can reach.
At the beginning of training, directly applying an extreme token budget
can produce highly degraded rollouts, making the visited states
far from the full-token policy. In other words, the lower the token budget is, more difficult the training task becomes. Motivated by previous work~\citep{bengio2009curriculum}, we introduce a budget-level curriculum learning strategy, to progressively increase the task difficulty during training. Specifically, let $s_{tol}$ be the total training steps and $s$ still be the current update, we define a three-stage curriculum process divided by $s_1$ and $s_2$, as Figure~\ref{fig:token} shows ($s_1$=14, $s_2$=100 empirically),
\begin{equation}
r(s)=
\begin{cases}
(2\alpha+1)\beta,
& 0\le s < s_1, \\[2mm]
\beta+\alpha\beta
\left[
1+\cos\left(
\pi\frac{s-s_1}{s_2-s_1}
\right)
\right],
& s_1\le s < s_2, \\[2mm]
\beta,
& s_2\le s\le s_{tol} ,
\end{cases}
\label{eq:curriculum}
\end{equation}
where $r(s)$ is the dynamic budget changed together with $s$ and $\alpha$ is a hyper-parameter 
giving the starting point of the learning. Here $\alpha$ is set to 2 empirically. $\beta$ is the final keep ratio.  In each training step, the number $N_{stu}$ of visual tokens student policy keep is $\mathrm{Round}(r(s)*N)$.

The first stage provides the student with more visual
evidence at a higher budget to establish useful on-policy trajectories. The second stage
progressively and smoothly shifts the visited state distribution towards that induced by
the target compression level, while the final stage adapts the student directly
at the target budget $\beta$. The complete LT-OPD objective at update $s$ is
therefore simply:
\begin{equation}
\mathcal{L}_{\mathrm{LT\text{-}OPD}}(\theta;s)
=
\widetilde{\mathcal{L}}_{\mathrm{OPD}}^{(s)}(\theta;r(s)).
\end{equation}

\subsection{Training Data Construction}
\label{sec:training_data}
We want the student model can see data from multiple sources, and then the supervision on student-generated trajectories can benefit from data diversity, and improve generalizability of the trained compressed model. Therefore, we specifically construct a training set, LT-14K, from multiple data sources. The training set contains 14,000 single-image questions from sources like LlaVA-Instruct-665K~\citep{liu2023visual}, OneThinker~\citep{feng2026onethinker}, PixMo~\citep{deitke2025molmo} and Vision-OPD~\citep{yuan2026vision,wei2026zooming} training data. Detailed distributions are in Appendix~\ref{app:data}. 

\vspace{-2mm}
\section{Experiments}
\vspace{-2mm}
\subsection{Experimental Settings.} 
\label{sec:settings}
\paragraph{Model and baselines.} We apply our methods on an advanced MLLM, Qwen-3.5-4B. We apply two kinds of baselines. For training-free baselines, we choose previous SOTA (state-of-the-art) methods like FastV~\citep{chen2024image}, DivPrune~\citep{alvar2025divprune}, DART~\citep{wen2025stop}, CDPruner~\citep{zhang2025beyond}, HiPrune~\citep{liu2026hiprune} and ZOO-Prune~\citep{kim2026zoo}, and set 20\%, 15\%, 10\% and 5\% keeping budget. For training-based ones, we choose vanilla finetuning (same token reduction method as LT-OPD), LlaVA-Mini~\cite{zhang2025llava}, EPIC~\cite{wen2025efficient} and LearnPruner~\citep{takezoe2026learnpruner}, and set the keeping budget of them all to 5\%. Here ``\textit{LlaVA-Mini}'' refers to the all fusion mechanism and training methods in their paper, which we reproduce on Qwen3.5-4B~\citep{qwen3.5}. For fair comparison, we all keep 5\% budget for them. We also apply another two MLLMs, GLM-4.6V-9B~\citep{hong2025glm} and LlaVA-OneVision (OV)-1.5-4B~\citep{an2025llava}  for cross-model evaluation.                        

\textbf{Basic settings.} Before trainining, we choose CDPruner~\citep{zhang2025beyond}, a token pruning method, based on conditional diversity, to reduce visual tokens of the model. It is easy for implementation. We train our model and training-based comparing methods on 8$\times$NVIDIA A100 80G GPUs. Evaluations are done on one same GPU. Detailed settings can be found in Appendix~\ref{app:implementation},~\ref{app:training_prompts}.

\begin{table}[t]
    \centering
    \caption{Main results of different visual token reduction methods on {\setlength{\fboxsep}{1pt}\colorbox{red!10}{Qwen3.5-4B}}. We \textbf{bold} the best and \underline{underline} the second best scores under \textit{5\% budget}. {\setlength{\fboxsep}{1pt}\colorbox{orange!10}{Orange}} lines are our results. ``Avg. Retain Ratio'' means average retain ratio compared to upper bound.}
    \label{tab:baseline}
    \setlength{\tabcolsep}{4pt}
    \resizebox{\textwidth}{!}{%
    \renewcommand{\arraystretch}{1.15}
    \begin{tabular}{lcccccccccc}
        \toprule
        \textbf{Method}
        & \textbf{V$^*$Bench}
        & \textbf{HR-4K}
        & \textbf{GQA}
        & \textbf{MMMU}
        & \textbf{MMB}
        & \textbf{MME}
        & \textbf{POPE}
        & \textbf{TVQA}
        & \textbf{OCRB}
        & \textbf{Avg. Retain Ratio (\%)} \\
        \midrule

        \rowcolor{gray!15}
        \multicolumn{11}{c}{\textit{Full Visual-Token Models (100\% Retention)}} \\
        Mimo-VL-RL & 83.3 & 73.5 & 55.9 & 66.7 & 84.4
        & 1,814.7 & 88.6 & 76.8 & 866 & - \\
        MiniCPM-V-4.5 & 70.7 & 69.6  & 61.0 & 67.7 & 84.2
        & 2,360.6 & 87.9 & 80.7 & 890 & - \\
        Qwen3-VL-Instruct-8B & 84.8 & 79.6  & 61.6 & 69.6 & 84.5
        & 2,387.4 & 89.0 & 81.7 & 819 & - \\
        \rowcolor{red!10} Qwen3.5-4B (\textit{Upper Bound})
        & 84.3 & 84.4 & 60.7  & 77.6 &  86.4
        & 2,300.8 & 89.3 & 81.2 & 850 & 100\% \\

        \midrule

        \rowcolor{gray!15}
        \multicolumn{11}{c}{\textbf{Training-Free}} \\

        \rowcolor{gray!10}
        \multicolumn{11}{c}{\textit{Retain 20\% Visual Tokens}} \\

        FastV{\footnotesize\texttt{(ECCV24)}}
        & 80.1 & 72.8 & 50.0 & 45.2 & 76.2
        & 1,729.4 & 77.7 & 47.2 & 468 & 76.2\% \\

        DivPrune{\footnotesize\texttt{(CVPR25)}}
        & 80.6 & 76.5  & 57.4 & 43.1 & 78.6
        & 1,938.8 & 88.7 & 70.3 & 579  & 85.1\% \\

        DART{\footnotesize\texttt{(EMNLP25)}}
        & 67.0 & 70.0  & 54.3 & 40.8 & 73.8
        & 1,909.8 & 84.8 & 73.8 & 579  & 80.8\% \\

        CDPruner{\footnotesize\texttt{(NeurIPS25)}}
        & 69.6 & 71.8  & 54.5 & 47.6 & 82.0
        & 1,949.2 & 80.0 & 63.7 & 463  & 80.1\% \\

        HiPrune{\footnotesize\texttt{(ACL26)}}
        & 71.7 & 72.1 & 55.4 & 48.7 & 81.6
        & 1,943.9 & 81.6 & 64.2 & 384 & 79.9\% \\

        ZOO-Prune{\footnotesize\texttt{(CVPR26)}}
        & 78.5 & 75.9  & 58.0 & 39.6 & 75.9
        & 2,009.4 & 89.2 & 56.0 & 556 & 82.1\% \\

        \rowcolor{gray!10}
        \multicolumn{11}{c}{\textit{Retain 15\% Visual Tokens}} \\

        FastV{\footnotesize\texttt{(ECCV24)}}
        & 78.0 & 72.0 & 47.1  & 46.1 & 73.0
        & 1,568.5 & 73.9 & 39.2 & 393 & 71.6\% \\

         DivPrune{\footnotesize\texttt{(CVPR25)}}
        & 78.5 & 76.6  & 56.1 & 42.0 & 78.0
        & 1,889.4 & 87.3 & 66.6 & 542  & 82.9\% \\

        DART{\footnotesize\texttt{(EMNLP25)}}
        & 62.8 & 68.0  & 52.7 & 41.1 & 72.2
        & 1,872.8 & 83.1 & 70.2 & 541  & 78.1\% \\

        CDPruner{\footnotesize\texttt{(NeurIPS25)}}
        & 64.9 & 70.6  & 53.2 & 48.0 & 80.6
        & 1,845.3 & 77.6 & 59.6 & 419  & 77.0\% \\

        HiPrune{\footnotesize\texttt{(ACL26)}}
        & 68.1 & 72.0  &  52.3 & 48.4 & 78.6
        & 1,750.8 & 78.6 & 55.0 & 279 & 74.5\% \\

        ZOO-Prune{\footnotesize\texttt{(CVPR26)}}
        & 78.0 & 75.3  & 56.5 & 42.9 & 74.7
        & 2,000.1 & 88.8 & 49.8 & 503 & 80.4\% \\

        \rowcolor{gray!10}
        \multicolumn{11}{c}{\textit{Retain 10\% Visual Tokens}} \\

        FastV{\footnotesize\texttt{(ECCV24)}}
        & 73.8 &  72.0 & 43.7 &  46.0 & 69.1
        & 1,452.4 & 69.3 & 35.1 & 305 & 67.1\% \\

        DivPrune{\footnotesize\texttt{(CVPR25)}}
        & 74.9 & 74.5  & 54.6 & 42.6 & 77.8
        & 1,873.8 & 86.0 & 60.0 & 496  & 80.2\% \\

        DART{\footnotesize\texttt{(EMNLP25)}}
        & 59.2 & 61.9  & 51.1 & 41.2 & 72.1
        & 1,803.8 & 81.8 & 63.5 & 505  & 74.7\% \\

        CDPruner{\footnotesize\texttt{(NeurIPS25)}}
        & 57.6 & 69.6 & 51.6 & 46.7 & 78.4
        & 1,742.2 & 74.3 & 53.9 & 415 & 73.4\% \\

        HiPrune{\footnotesize\texttt{(ACL26)}}
        & 63.9 &  70.2 & 49.0 & 47.6 & 75.6
        & 1,471.0 & 72.2 & 40.3 & 202 & 67.4\% \\

        ZOO-Prune{\footnotesize\texttt{(CVPR26)}}
        & 77.5 & 73.6  & 55.2 & 40.1 & 74.2
        & 1,949.4 & 87.8 & 40.1 & 469 & 77.2\% \\

        \rowcolor{gray!10}
        \multicolumn{11}{c}{\textit{Retain 5\% Visual Tokens}} \\

        FastV{\footnotesize\texttt{(ECCV24)}}
        & 63.4 & 66.5  & 39.8 & 45.8 & 61.2
        & 1,193.2 & 61.9 & 31.0 & 187 & 59.0\% \\

        DivPrune{\footnotesize\texttt{(CVPR25)}}
        & 70.7 & \textbf{74.4}  & 54.6 & 43.7 & 77.2
        & 1,834.8 & 86.0 & 48.6 & 467  & 77.6\% \\

        DART{\footnotesize\texttt{(EMNLP25)}}
        & 52.9 & 57.6  & 51.1 & 42.8 & 71.8
        & 1,771.3 & 81.7 & 52.3 & 477  & 71.4\% \\

        CDPruner{\footnotesize\texttt{(NeurIPS25)}}
        & 52.9 & 68.4 & 50.4 & 45.7 & 74.0
        & 1,531.8 & 70.2 & 43.2 & 404 & 68.6\% \\

        HiPrune{\footnotesize\texttt{(ACL26)}}
        & 58.7 & 68.3  & 45.4 & 47.2 & 71.2
        & 1,372.9 & 63.7 & 25.5 & 107 & 60.4\% \\

        ZOO-Prune{\footnotesize\texttt{(CVPR26)}}
        & 70.7 & 71.3  & 54.1 & 38.3 & 74.1
        & 1,915.4 & 85.0 & 26.8 & 431 & 72.7\% \\

        \midrule

        \rowcolor{gray!15}
        \multicolumn{11}{c}{\textbf{Training-Based (5\% Visual Tokens)}} \\

        \textit{LLaVA-Mini}{\footnotesize\texttt{(ICLR25)}}
        & 62.8 & 70.1  & 47.8 & 47.2 & 78.2
        & 1,879.6 & 78.0 & 45.2 & 370 & 72.9\% \\

        EPIC{\footnotesize\texttt{(NeurIPS25)}}
        & 68.6 & 72.8  & 43.1 & 47.3 & 77.8
        & 1,781.9 & 81.1 & 40.5 & 470 & 73.7\% \\

        LearnPruner{\footnotesize\texttt{(ICLR26)}}
        & 66.0 & 61.4 & 48.7 & 47.1 & 73.3
        & 1,534.9 &  73.9 & 35.0 & 403 & 68.5\% \\

        5\% Token + SFT & 53.9  & 66.2 & 48.8 & 47.8 & 77.4
        & 1,741.7 & 72.5 & 51.6 & 330 & 70.4\% \\

        5\% Token + curriculum + SFT
        & 61.8 & 71.8 & 49.2 & 47.4 & 77.7
        & 1,715.2 & 72.9 & 53.7 & 380 & 73.1\% \\

        5\% Token + curriculum + fixed-prefix JSD
        & 65.5 & 69.3 & 52.3 & 48.4 & 77.7
        & 1,839.0 & 83.5 & 53.2 & 520 & 77.6\% \\

        \rowcolor{orange!10}
        \textbf{LT-OPD} \footnotesize{(w/o curriculum)}
        & \underline{74.3} & 72.1  & \underline{54.7} & \underline{49.6} & \textbf{78.9}
        &  \underline{1,995.5}   & \textbf{86.5} & \textbf{56.3} & \underline{524} & \underline{81.5\%} \\
        \rowcolor{orange!10}
        \textbf{LT-OPD}
        & \textbf{74.9} & \underline{73.0}  & \textbf{54.8} & \textbf{50.0} & \textbf{78.9}
        &   \textbf{2,077.0}   & \underline{86.3} & \underline{56.0} & \textbf{535}  & \textbf{82.3\%} \\
    \end{tabular}%
    }
    \vspace{-8mm}
\end{table}

\textbf{Datasets and benchmarks.} All the training-based methods are trained on the constructed LT-14K for fair comparison. For evaluations, we choose four kinds of benchmarks: (1) perceptual-heavy benchmarks, V$^*$Bench~\citep{wu2024v} and HR-Bench 4K (HR-4K)~\citep{wang2025divide}, which require fine-grained visual information; (2) general VQA benchmarks, GQA~\citep{hudson2019gqa}, MMMU~\citep{yue2024mmmu}, MMB (en)~\citep{liu2024mmbench} and MME~\citep{fu2026mme} (3) visual hallucination benchmarks, POPE~\citep{li2023evaluating}; (4) OCR benchmarks, TextVQA (TVQA)~\citep{singh2019towards} and OCRBench (OCRB)~\citep{liu2024ocrbench}. Evaluation prompts are in Appendix~\ref{app:evaluation}.

\textbf{Evaluation metrics.} We follow the original metrics for judgment of the chosen benchmarks. Notably, we also calculate the average retain ratio compared to upper bound, to intuitively quantify how much of the full-token model's capability is preserved under token reduction: $\text{average \ retain \ ratio}
=
\frac{1}{M}
\sum_{i=1}^{M}
\frac{score_i}{score_i^{\mathrm{full}}}
\times 100\%,$ where $M$ benchmarks are chosen and $score_i$, $score_i^{\mathrm{full}}$ mean the scores of compressed and full-token model.

\subsection{Main results}
\textbf{Effectiveness of LT-OPD.} Table~\ref{tab:baseline} compares LT-OPD with both training-free and training-based visual token reduction methods on Qwen3.5-4B. 
Under the extreme 5\% visual-token budget, LT-OPD achieves an average retain ratio of 82.3\%, outperforming the strongest training-free baseline, DivPrune, by 4.7 percentage points (77.6\%), and training-based baseline, EPIC by 8.6 points (73.7\%). 
The improvement is broad rather than benchmark-specific: LT-OPD achieves the best result among prior methods keeping 5\% visual tokens on \textbf{8 out of 9 benchmarks}, including all categories. 
The gains are particularly clear on V$^*$Bench (74.9 vs. 70.7), MME (2,077.0 vs. 1,915.4), and OCRBench (535 vs. 477), where the latter numbers denote the best competing results under the same budget. Additionally, the chosen visual token reduction strategy for LT-OPD, CDPruner, achieves an average retain ratio of only 68.6\% under 5\% budget, much lower than LT-OPD. This suggests that LT-OPD does not rely on a ``best'' token reduction method, but proper on-policy learning actually works.

More notably, with only 5\% visual tokens, LT-OPD still outperforms all training-free methods retaining 10\% tokens in average retain ratio (82.3\% vs. at most 80.2\%), and nearly matches the best result obtained with a 15\% or even 20\% budget.
These results demonstrate that with effective training, our method can recover substantially more model capability even under extreme low token budget than relying on token selection alone or applying simple training.

\textbf{Ablation Studies.}
We ablate the proposed OPD training and curriculum learning strategy, reported in the last 4 lines of Table~\ref{tab:baseline}. It is obvious that both components can improve training performance, as curriculum learning increases average retain ratio of simply SFT by 2.7 percentage points, and OPD-only training reaches 81.5\% average retain ratio. Based on this, full-version LT-OPD achieves the best performance (82.3\%) and clearly improves model performance across all benchmarks. More notably, training with curriculum and JSD with fixed prefix (knowledge distillation) reaches a score of 77.6, still lower than our results, suggesting that on-policy learning really benefits LT-OPD. 

We also test different training objectives, with explanations and results displayed in Appendix~\ref{app:objectives}. While all training objectives show promising performance, our objective, JSD, is better on perception-heavy benchmarks. Forward/Reverse KL, however, achieve better scores on OCR benchmarks. The whole ablation study suggests the effectiveness and stability of our full design.

\vspace{-2mm}
\subsection{Different Model Scales.}
\vspace{-2mm}
\label{sec:model_scale}
To examine whether LT-OPD remains effective across different model scales, we evaluate it on the 4B and 9B variants of Qwen3.5 under the same 5\% visual-token budget. As shown in Table~\ref{tab:model_scale}, LT-OPD consistently improves both model sizes across all evaluated benchmarks. For Qwen3.5-4B, the average retained performance increases from 68.6\% to 82.3\%, while for Qwen3.5-9B it improves from 71.1\% to 81.4\%. Notably, although the larger 9B model provides a stronger compressed baseline, LT-OPD still yields substantial gains, particularly on V$^*$Bench, HR-Bench, and so on. These results indicate that the effectiveness of LT-OPD is not tied to a particular model scale. The benefit of our method persists as the underlying MLLM scales from 4B to 9B parameters.

\begin{table}[t]
    \centering
    \caption{Different model scales. We apply LT-OPD on 4B and 9B version. ``Avg.'' means average retain ratio compared to upper bound (4B or 9B full), the same as Table~\ref{tab:baseline}.}
    \label{tab:model_scale}
    \resizebox{\textwidth}{!}{%
        \renewcommand{\arraystretch}{1.15}
        \begin{tabular}{lcccccccccc}
            \toprule
            \textbf{Method}
            & \textbf{V$^*$Bench}
            & \textbf{HR-4K}
            & \textbf{GQA}
            & \textbf{MMMU}
            & \textbf{MMB}
            & \textbf{MME}
            & \textbf{POPE}
            & \textbf{TextVQA}
            & \textbf{OCRB}
            & \textbf{Avg.} \\
            \midrule




            Qwen3.5-4B (5\% visual token)
            & 52.9 & 68.4 & 50.4 & 45.7 & 74.0
            & 1,531.8 & 70.2 & 43.2 & 404 & 68.6\% \\

            \rowcolor{orange!10}
            \textbf{+ LT-OPD}
                & \textbf{74.9} & \textbf{73.0} & \textbf{54.8}
            & \textbf{50.0} & \textbf{78.9} & \textbf{2,077.0}
            & \textbf{86.3} & \textbf{56.0} & \textbf{535}
            & \textbf{82.3\%} \\

            Qwen3.5-9B (5\% visual token)
            & 54.5 & 59.6 & 49.4 & 51.7 & 77.7
            & 1,809.4 & 82.3 & 46.6 & 430 & 71.1\% \\

            \rowcolor{orange!10}
            \textbf{+ LT-OPD}
            & \textbf{70.7} & \textbf{68.1} & \textbf{53.6}
            & \textbf{52.9} & \textbf{80.5} & \textbf{2,147.9}
            & \textbf{86.2} & \textbf{61.5} & \textbf{539}
            & \textbf{81.4\%} \\
            \bottomrule
        \end{tabular}%
        }
        \vspace{-4mm}
\end{table}

\vspace{-2mm}
\subsection{Cross-Model Performance.}
\vspace{-2mm}
\begin{table}[t]
    \centering
    \caption{Performance of LT-OPD on different kinds of MLLMs. ``Avg.'' means average retain ratio compared to upper bound (full version of models), the same as Table~\ref{tab:baseline}.}
    \label{tab:cross_model}
    \resizebox{\textwidth}{!}{%
    \renewcommand{\arraystretch}{1.15}
    \begin{tabular}{lcccccccccc}
        \toprule
        \textbf{Method}
        & \textbf{V$^*$Bench}
        & \textbf{HR-4K}
        & \textbf{GQA}
        & \textbf{MMMU}
        & \textbf{MMB}
        & \textbf{MME}
        & \textbf{POPE}
        & \textbf{TextVQA}
        & \textbf{OCRB}
        & \textbf{Avg.} \\
        \midrule



        Qwen3.5-4B (5\% visual token)
            & 52.9 & 68.4 & 50.4 & 45.7 & 74.0
            & 1,531.8 & 70.2 & 43.2 & 404 & 68.6\% \\

        \rowcolor{orange!10}
        \textbf{+ LT-OPD}
            & \textbf{74.9} & \textbf{73.0} & \textbf{54.8}
            & \textbf{50.0} & \textbf{78.9} & \textbf{2,077.0}
            & \textbf{86.3} & \textbf{56.0} & \textbf{535}
            & \textbf{82.3\%} \\

        GLM-4.6V-9B + (5\% visual token)  &45.0  &60.5 &46.8  &42.3  &69.2 
        &1,646.0 &76.3  &56.1  &346 & 73.8\% \\
        
       \rowcolor{orange!10} \textbf{+ LT-OPD} (5\% visual token)  & \textbf{51.3} & \textbf{63.4} & \textbf{53.8} & \textbf{48.3} & \textbf{83.4} & \textbf{2,146.3} & \textbf{83.3} & \textbf{64.8} & \textbf{497} & \textbf{86.3\%} \\

        LlaVA-OV-1.5-4B + (5\% visual token)
        &64.9   &51.9  &50.9  &47.8  &64.6 
        &1,532.1 &81.9  &38.6  &87 & 71.8\% \\

       \rowcolor{orange!10} \textbf{+ LT-OPD} (5\% visual token)
        & 64.9  & \textbf{54.0}   & \textbf{55.9}  & \textbf{49.9}  & \textbf{78.5}
        & \textbf{1,940.5} & \textbf{87.4}  & \textbf{46.8}   & \textbf{430} & \textbf{84.1\%} \\

        \bottomrule
    \end{tabular}%
    }
    \vspace{-3mm}
\end{table}
We apply LT-OPD on another two advanced MLLMs, GLM-4.6V-9B and LlaVA-OV-1.5-4B. Table~\ref{tab:cross_model} reports the results. LT-OPD consistently improves the performance of both models under the same 5\% visual-token budget, increasing the average retained performance from 73.8\% to 86.3\% on GLM-4.6V-9B and from 71.8\% to 84.1\% on LLaVA-OV-1.5-4B. The gains are observed across a broad range of benchmarks, with particularly clear improvements like MMBench and MME. These results demonstrate that the effectiveness of LT-OPD is not specific to the Qwen family, but transfers consistently to different MLLM backbones, suggesting that on-policy self-distillation provides a broadly effective training signal for models under extremely limited visual-token budgets.

\subsection{Efficiency}
\vspace{-2mm}
\textbf{Inference efficiency.} Table~\ref{tab:efficiency} summarizes the inference efficiency of LT-OPD (evaluation details in Appendix~\ref{app:additional_evaluation}). By retaining only 5\% visual tokens, LT-OPD reduces the KV cache by 85.2\%, and both overall and prefill FLOPs by 85.4\%, while introducing no additional inference overhead. It achieves the same computational cost as HiPrune and requires substantially fewer FLOPs than EPIC (0.146 vs. 0.200), demonstrating that the performance gains of LT-OPD do not compromise the efficiency of the compressed model. Additionally, the end-to-end latency improves moderately from 1331.5 to 1293.7 ms ($1.029\times$), consistent with prior observations~\citep{wen2025efficient,zamini2026delta} that aggressive visual-token reduction primarily benefits prefill computation, while  autoregressive decoding time is dominated by per-token computation and memory access.

\textbf{Training efficiency.} Our training method trains on only 14K training data. The training loop ran for 15.1h on eight GPUs. Considering the study of training time in the original paper~\cite{wen2025efficient} that 12.2-48 hours are used for training on the same devices (EPIC, LlaVA-Mini, etc.), our training cost is acceptable under the scenario of training-based visual token reduction.

\begin{table}[t]
\centering
\caption{Efficiency of LT-OPD. $\Delta$ means comparing with raw Qwen3.5-4B. On the 5\% budget, LT-OPD can reduce prefilling cost a lot. Details are in Appendix~\ref{app:additional_evaluation}.}
\label{tab:efficiency}
\renewcommand{\arraystretch}{1.15}
\resizebox{\linewidth}{!}{
\begin{tabular}{lcccccccc}
\toprule
\textbf{Method}
& \multicolumn{2}{c}{\textbf{KV Cache (MB)} $\downarrow$}
& \multicolumn{2}{c}{\textbf{FLOPs Ratio} $\downarrow$}
& \multicolumn{2}{c}{\textbf{Prefill FLOPs (T)} $\downarrow$}
& \multicolumn{2}{c}{\textbf{End-to-End Latency (ms/sample)} $\downarrow$}\\
\cmidrule(lr){2-3}\cmidrule(lr){4-5}\cmidrule(lr){6-7}\cmidrule(lr){8-9}
& \textbf{Value} & $\Delta$
& \textbf{Value} & $\Delta$
& \textbf{Value} & $\Delta$
& \textbf{Value} & \textbf{Speedup} $\uparrow$\\
\midrule


    Qwen3.5-4B (full visual token)
    & 26.59 & -- & 1.000 & -- & 6.002 & -- & 1331.510 & $1.000\times$ \\

    \textbf{+ HiPrune} (5\% visual token)
    & 3.93 & 85.23\% & 0.146 & 85.38\% & 0.878 & 85.38\% & 1308.029 & $1.018\times$ \\

    \textbf{+ EPIC} (5\% visual token)
    & 3.93 & 85.23\% & 0.200 & 80.03\% & 1.198 & 80.03\% & 1296.922 & $1.027\times$ \\

    \textbf{+ LT-OPD} (5\% visual token)
    & 3.93 & 85.23\% & 0.146 & 85.38\% & 0.878 & 85.38\% & 1293.691 & $1.029\times$ \\

    \bottomrule
\end{tabular}
}
\vspace{-4mm}
\end{table}

\begin{table}[t]
\centering
\caption{Compared with advanced RL algorithms. ``Avg.'' means average retain ratio compared to upper bound, the same as Table~\ref{tab:baseline}.}
\label{tab:rl}
\resizebox{\textwidth}{!}{%
\renewcommand{\arraystretch}{1.15}
\begin{tabular}{lcccccccccc}
\toprule
\textbf{Method}
& \textbf{V$^*$Bench}
& \textbf{HR-4K}
& \textbf{GQA}
& \textbf{MMMU}
& \textbf{MMB}
& \textbf{MME}
& \textbf{POPE}
& \textbf{TextVQA}
& \textbf{OCRB}
& \textbf{Avg.} \\
\midrule



Qwen3.5-4B (5\% visual token)
& 52.9 & 68.4 & 50.4 & 45.7 & 74.0
& 1,531.8 & 70.2 & 43.2 & 404 & 68.6\% \\

+ GRPO
& 56.5 & 66.8 & 53.9
& 49.0 & 77.1 & 1,958.1
& 82.7 & 49.4 & 479
& 75.8\% \\

+ GSPO
& 58.6 & 63.9
& 53.5 & 49.9 & 78.3
& 2,072.3
& 85.7 & 50.6 & 467
& 76.8\% \\

+ DAPO
& 60.7 & 67.9
& 54.4 & 49.6 & 78.0
& 2,037.7
& 85.8 & 52.1 & 497
& 78.2\% \\

\rowcolor{red!10}
\textbf{+ LT-OPD}
& \textbf{74.9} & \textbf{73.0} & \textbf{54.8}
& \textbf{50.0} & \textbf{78.9}
& \textbf{2,077.0} & \textbf{86.3}
& \textbf{56.0} & \textbf{535}
& \textbf{82.3\%} \\
\bottomrule
\end{tabular}%
}
\vspace{-5mm}
\end{table}

\begin{table}[t]
    \centering
    \caption{Different teacher regularization strategies on Qwen3.5-4B. ``Avg.'' means average retain ratio compared to upper bound, the same as Table~\ref{tab:baseline}}
    \label{tab:regularization}
    \resizebox{\textwidth}{!}{%
        \renewcommand{\arraystretch}{1.15}
        \begin{tabular}{lcccccccccc}
            \toprule
            \textbf{Teacher}
            & \textbf{V$^*$Bench}
            & \textbf{HR-4K}
            & \textbf{GQA}
            & \textbf{MMMU}
            & \textbf{MMB}
            & \textbf{MME}
            & \textbf{POPE}
            & \textbf{TextVQA}
            & \textbf{OCRB}
            & \textbf{Avg.} \\
            \midrule




            \rowcolor{orange!7} Initial Policy
            & 74.9 & 73.0 & 54.8
            & 50.0 & 78.9 & 2,077.0
            & 86.3 & 56.0 & 535
            & 82.3\% \\
            \rowcolor{green!7} Trust-Region Regularization
            & 74.3 & 72.1 & 52.9 & 49.0 & 79.1
            & 2,115.0 & 86.4 & 55.4 & 518 & 81.5\% \\
            \rowcolor{blue!7} EMA 
            & 75.4 & 73.4 & 54.1 & 50.0 & 77.6 
            & 2,026.6 & 84.9 & 55.7 & 511 & 81.3\% \\
            \bottomrule
        \end{tabular}%
    }
    \vspace{-4mm}
\end{table}
\begin{figure}[htbp]
  \centering
  \includegraphics[width=\linewidth]{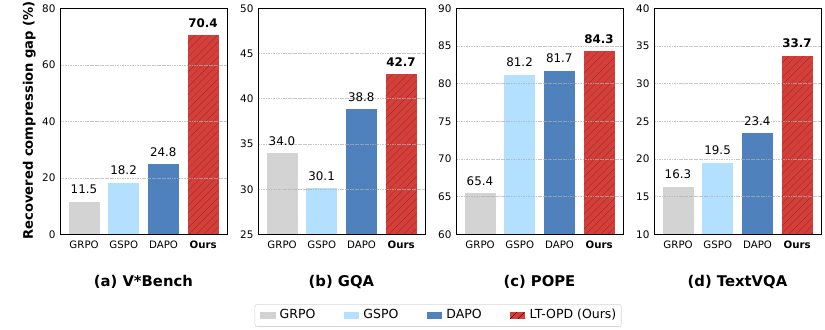}
  \vspace{-4mm}
  \caption{\textbf{Recovery at 5\% visual-token retention.} We report the scores of LT-OPD and compared RL algorithms. LT-OPD achieves the best recovery performances.}
  \label{fig:compression_gap_recovery}
  \vspace{-4mm}
\end{figure}

\vspace{-2mm}
\subsection{Compared with RL algorithms}
\vspace{-2mm}
\label{sec:rl}
We compare our method with three advanced RL algorithms: GRPO~\citep{shao2024deepseekmath}, GSPO~\citep{zheng2025gspo} and DAPO~\citep{yu2026dapo}. As shown in Table~\ref{tab:rl}, directly applying these general-purpose RL algorithms substantially improves the 5\% visual-token baseline, increasing the average retain ratio from 68.6\% to 75.8\%--78.2\%. Nevertheless, LT-OPD consistently outperforms all three alternatives across all benchmarks and further raises the average retain ratio to 82.3\%. In particular, compared with the strongest RL baseline DAPO, LT-OPD achieves notable gains on V$^*$Bench, MME, and TextVQA. These results suggest that, under an extremely limited visual-token budget, distilling the model's on-policy knowledge provides a more effective learning signal than optimizing with general-purpose RL objectives, enabling the compressed model to better recover the capabilities of its full-token counterpart.

\subsection{Extensive Studies}
\vspace{-2mm}
\textbf{Training Strategies.} Table~\ref{tab:regularization} reports performance with different training strategies of OPSD, mentioned in~\citep{zhao2026self,yuan2026vision}. Details are in Appendix~\ref{app:strategy}. LT-OPD primarily applies the frozen initial policy as the teacher. We discover that LT-OPD keeps stable and good performance across all benchmarks, with the scores and average retain ratio do not change much (81.3\%-82.3\%). This study supports the robustness of our method even using different regularization methods.

\textbf{Compression-gap recovery.} We want to explore whether our training algorithms better help the student policy to ``recover'' performance lost to visual token pruning. Thus, we define a new metric, recovery, calculated by $100\,(S_{\mathrm{method}}-S_{\mathrm{pruned}})/(S_{\mathrm{full}}-S_{\mathrm{pruned}})$. $S_{\{\mathrm{full},\ \mathrm{pruned}, \ \mathrm{method}\}}$ denotes original scores and scores before and after training. Larger recovery represents that a better policy is provided. Figure~\ref{fig:compression_gap_recovery} compares the scores between LT-OPD and RL algorithms.  LT-OPD recovers the most across four evaluated benchmarks, especially much higher (over 45 points) than RL methods on V$^*$Bench. This comparison further supports the results in Section~\ref{sec:rl}.

\textbf{Results on multi-image benchmarks.} We apply our method and baselines on two multi-image benchmarks, MuirBench~\citep{wang2025muirbench} and MVBench~\citep{li2024mvbench}. Results are in Appendix~\ref{app:multi_image}. LT-OPD outperforms its training-free counterpart (CDPruner) and EPIC a lot (by over 12\% in average retain ratio). We may draw the conclusion that LT-OPD has the potential to be generalized to video understanding tasks, which lead one of the future directions~\citep{zhang2026deep}. 

\textbf{Case Studies.} Some case studies are in Appendix~\ref{app:cases}. These cases range from questions natural images to OCR questions. LT-OPD can do well even on cases needing fine-grained visual information understanding or small texts compared with baselines. This suggests that students trained with LT-OPD have strong potential across different perceptual tasks.

\section{Conclusion}
\vspace{-2mm}
\label{sec:conclusion}
Extreme visual-token reduction changes more than the amount of visual evidence available to an MLLM: it also changes the generation states that the compressed model encounters. Motivated by this observation, we introduce LT-OPD, which distills a full-token self-teacher directly on trajectories generated by the low-token student, and progressively reduces the visual-token budget for stable training. Extensive experiments across nine benchmarks, multiple model scales, and different MLLM families show that LT-OPD substantially recovers the capability lost under 5\% visual-token retention without sacrificing the efficiency benefits of compression. These results point to a broader direction for efficient MLLMs: under extreme compression, effective token selection and effective adaptation should be considered jointly rather than in isolation.

\subsection*{AI use statement}

In this work, we used generative AI tools for assist with translation.
We have not used generative AI tools for things like generating data and supporting qualitative and thematic data analysis,
and the rest of the required disclosure tasks like mathematical proofs are not applicable to this work.
Additionally, we used generative AI tools for polishing our language. We have reviewed all AI-assisted work. We
checked LLM-generated research ideas for potential plagiarism through a manual
literature survey. LLM-generated code was verified and tested for correctness
by all of the authors. We take responsibility for the final content of this work,
including text, claims or artifacts produced with the aid of generative AI.

\subsection*{Ethics statement}

Our paper does not include any human
subjects, practices to data set releases, potentially harmful insights,
methodologies and applications, potential conflicts of interest,
discrimination/bias/fairness concerns, privacy and security issues, legal
compliance, and research integrity issues.

\subsection*{Reproducibility statement}

The training dataset details and benchmark details are in Section~\ref{sec:training_data}, Section~\ref{sec:settings} and Appendix~\ref{app:data}. Training details are in Appendix~\ref{app:implementation}. We also include the prompts in Appendix~\ref{app:training_prompts} and Appendix~\ref{app:evaluation}. Evaluation protocols on efficiency are in Appendix~\ref{app:additional_evaluation}. 



\bibliography{iclr2027_conference}

@inproceedings{zhang2025llava,
  title={Llava-mini: Efficient image and video large multimodal models with one vision token},
  author={Zhang, Shaolei and Fang, Qingkai and Zhe, Yang and Feng, Yang},
  booktitle={ICLR},
  year={2025}
}

@inproceedings{takezoe2026learnpruner,
  title={Learnpruner: Rethinking attention-based token pruning in vision language models},
  author={Takezoe, Rinyoichi and Li, Yaqian and Bo, Zi-Hao and Hou, Anzhou and Guang, Mo and Long, Kaiwen},
  booktitle={ICLR},
  year={2026}
}

@inproceedings{chen2024image,
  title={An image is worth 1/2 tokens after layer 2: Plug-and-play inference acceleration for large vision-language models},
  author={Chen, Liang and Zhao, Haozhe and Liu, Tianyu and Bai, Shuai and Lin, Junyang and Zhou, Chang and Chang, Baobao},
  booktitle={ECCV},
  year={2024}
}

@inproceedings{alvar2025divprune,
  title={Divprune: Diversity-based visual token pruning for large multimodal models},
  author={Alvar, Saeed Ranjbar and Singh, Gursimran and Akbari, Mohammad and Zhang, Yong},
  booktitle={CVPR},
  year={2025}
}

@inproceedings{wen2025stop,
  title={Stop looking for “important tokens” in multimodal language models: Duplication matters more},
  author={Wen, Zichen and Gao, Yifeng and Wang, Shaobo and Zhang, Junyuan and Zhang, Qintong and Li, Weijia and He, Conghui and Zhang, Linfeng},
  booktitle={EMNLP},
  year={2025}
}

@inproceedings{liu2026hiprune,
  title={HiPrune: Hierarchical Attention for Efficient Token Pruning in Vision-Language Models},
  author={Liu, Jizhihui and Zhu, Guangdao and Du, Feiyi and Lian, Niu and Li, Jun and Chen, Bin and Guan, Weili and Wang, Yaowei},
  booktitle={Findings of the Association for Computational Linguistics: ACL 2026},
  year={2026}
}

@inproceedings{zhang2025beyond,
  title={Beyond attention or similarity: Maximizing conditional diversity for token pruning in mllms},
  author={Zhang, Qizhe and Liu, Mengzhen and Li, Lichen and Lu, Ming and Zhang, Yuan and Pan, Junwen and She, Qi and Zhang, Shanghang},
  booktitle={NeurIPS},
  year={2025}
}

@inproceedings{wen2025efficient,
  title={Efficient multi-modal large language models via progressive consistency distillation},
  author={Wen, Zichen and Wang, Shaobo and Zhou, Yufa and Zhang, Junyuan and Zhang, Qintong and Gao, Yifeng and Chen, Zhaorun and Wang, Bin and Li, Weijia and He, Conghui and others},
  booktitle={NeurIPS},
  year={2025}
}

@inproceedings{agarwal2024policy,
  title={On-policy distillation of language models: Learning from self-generated mistakes},
  author={Agarwal, Rishabh and Vieillard, Nino and Zhou, Yongchao and Stanczyk, Piotr and Ramos Garea, Sabela and Geist, Matthieu and Bachem, Olivier},
  booktitle={ICLR},
  year={2024}
}

@article{xu2026deepseek,
  title={Deepseek-v4: Towards highly efficient million-token context intelligence},
  author={Xu, Anyi and Lin, Bangcai and Xue, Bing and Wang, Bingxuan and Xu, Bingzheng and Wu, Bochao and Zhang, Bowei and Lin, Chaofan and Dong, Chen and Ling, Chenchen and others},
  journal={arXiv preprint arXiv:2606.19348},
  year={2026}
}

@article{hou2026uni,
  title={Uni-opd: Unifying on-policy distillation with a dual-perspective recipe},
  author={Hou, Wenjin and Peng, Shangpin and Wang, Weinong and Ruan, Zheng and Zhang, Yue and Zhou, Zhenglin and Gao, Mingqi and Chen, Yifei and Wang, Kaiqi and Yang, Hongming and others},
  journal={arXiv preprint arXiv:2605.03677},
  year={2026}
}

@article{li2026rethinking,
  title={Rethinking on-policy distillation of large language models: Phenomenology, mechanism, and recipe},
  author={Li, Yaxuan and Zuo, Yuxin and He, Bingxiang and Zhang, Jinqian and Xiao, Chaojun and Qian, Cheng and Yu, Tianyu and Gao, Huan-ang and Yang, Wenkai and Liu, Zhiyuan and others},
  journal={arXiv preprint arXiv:2604.13016},
  year={2026}
}

@inproceedings{xu2025speculative,
  title={Speculative knowledge distillation: Bridging the teacher-student gap through interleaved sampling},
  author={Xu, Wenda and Han, Rujun and Wang, Zifeng and Le, Long and Madeka, Dhruv and Li, Lei and Wang, William and Agarwal, Rishabh and Lee, Chen-Yu and Pfister, Tomas},
  booktitle={ICLR},
  year={2025}
}

@article{shao2024deepseekmath,
  title={Deepseekmath: Pushing the limits of mathematical reasoning in open language models},
  author={Shao, Zhihong and Wang, Peiyi and Zhu, Qihao and Xu, Runxin and Song, Junxiao and Bi, Xiao and Zhang, Haowei and Zhang, Mingchuan and Li, YK and Wu, Yang and others},
  journal={arXiv preprint arXiv:2402.03300},
  year={2024}
}

@inproceedings{yu2026dapo,
  title={Dapo: An open-source llm reinforcement learning system at scale},
  author={Yu, Qiying and Zhang, Zheng and Zhu, Ruofei and Yuan, Yufeng and Zuo, Xiaochen and Yue, Yu and Dai, Weinan and Fan, Tiantian and Liu, Gaohong and Liu, Lingjun and others},
  booktitle={NeurIPS},
  year={2025}
}

@article{zheng2025gspo,
      title={Group Sequence Policy Optimization}, 
      author={Chujie Zheng and Shixuan Liu and Mingze Li and Xiong-Hui Chen and Bowen Yu and Chang Gao and Kai Dang and Yuqiong Liu and Rui Men and An Yang and Jingren Zhou and Junyang Lin},
      journal={arXiv preprint arXiv:2507.18071},
      year={2025}
}

@article{li2026g,
  title={G$^2$TR: Generation-Guided Visual Token Reduction for Separate-Encoder Unified Multimodal Models},
  author={Li, Junxian and Liu, Kai and Ding, Zizhong and Wang, Zhixin and Chen, Zhikai and Pei, Renjing and Zhang, Yulun},
  journal={arXiv preprint arXiv:2605.12309},
  year={2026}
}

@inproceedings{zamini2026delta,
  title={Delta-LLaVA: Base-then-Specialize Alignment for Token-Efficient Vision-Language Models},
  author={Zamini, Mohamad and Shukla, Diksha},
  booktitle={WACV},
  year={2026}
}

@article{an2025llava,
  title={Llava-onevision-1.5: Fully open framework for democratized multimodal training},
  author={An, Xiang and Xie, Yin and Yang, Kaicheng and Zhang, Wenkang and Zhao, Xiuwei and Cheng, Zheng and Wang, Yirui and Xu, Songcen and Chen, Changrui and Zhu, Didi and others},
  journal={arXiv preprint arXiv:2509.23661},
  year={2025}
}

@article{hong2025glm,
  title={Glm-4.5 v and glm-4.1 v-thinking: Towards versatile multimodal reasoning with scalable reinforcement learning},
  author={Hong, Wenyi and Yu, Wenmeng and Gu, Xiaotao and Wang, Guo and Gan, Guobing and Tang, Haomiao and Cheng, Jiale and Qi, Ji and Ji, Junhui and Pan, Lihang and others},
  journal={arXiv preprint arXiv:2507.01006},
  year={2025}
}

@misc{qwen3.5,
    title  = {{Qwen3.5}: Towards Native Multimodal Agents},
    author = {{Qwen Team}},
    month  = {February},
    year   = {2026},
    url    = {https://qwen.ai/blog?id=qwen3.5}
}

@inproceedings{pmlr-v267-chu25c,
  title = 	 {{SFT} Memorizes, {RL} Generalizes: A Comparative Study of Foundation Model Post-training},
  author =       {Chu, Tianzhe and Zhai, Yuexiang and Yang, Jihan and Tong, Shengbang and Xie, Saining and Schuurmans, Dale and Le, Quoc V and Levine, Sergey and Ma, Yi},
  booktitle = 	 {ICML},
  year = 	 {2025}
}

@inproceedings{zhao2026self,
  title={Self-distilled reasoner: On-policy self-distillation for large language models},
  author={Zhao, Siyan and Xie, Zhihui and Liu, Mengchen and Huang, Jing and Pang, Guan and Chen, Feiyu and Grover, Aditya},
  booktitle={ICML},
  year={2026}
}

@article{yuan2026vision,
  title={Vision-opd: Learning to see fine details for multimodal llms via on-policy self-distillation},
  author={Yuan, Qianhao and Lou, Jie and Yu, Xing and Lin, Hongyu and Sun, Le and Han, Xianpei and Lu, Yaojie},
  journal={arXiv preprint arXiv:2605.18740},
  year={2026}
}

@inproceedings{zhang2025lmms,
  title={Lmms-eval: Reality check on the evaluation of large multimodal models},
  author={Zhang, Kaichen and Li, Bo and Zhang, Peiyuan and Pu, Fanyi and Cahyono, Joshua Adrian and Hu, Kairui and Liu, Shuai and Zhang, Yuanhan and Yang, Jingkang and Li, Chunyuan and others},
  booktitle={NAACL Findings},
  year={2025}
}

@inproceedings{duan2024vlmevalkit,
  title={Vlmevalkit: An open-source toolkit for evaluating large multi-modality models},
  author={Duan, Haodong and Yang, Junming and Qiao, Yuxuan and Fang, Xinyu and Chen, Lin and Liu, Yuan and Dong, Xiaoyi and Zang, Yuhang and Zhang, Pan and Wang, Jiaqi and others},
  booktitle={ACM MM},
  year={2024}
}

@inproceedings{wu2024v,
  title={V*: Guided visual search as a core mechanism in multimodal llms},
  author={Wu, Penghao and Xie, Saining},
  booktitle={CVPR},
  year={2024}
}

@inproceedings{wang2025divide,
  title={Divide, conquer and combine: A training-free framework for high-resolution image perception in multimodal large language models},
  author={Wang, Wenbin and Ding, Liang and Zeng, Minyan and Zhou, Xiabin and Shen, Li and Luo, Yong and Yu, Wei and Tao, Dacheng},
  booktitle={AAAI},
  year={2025}
}

@inproceedings{hudson2019gqa,
  title={Gqa: A new dataset for real-world visual reasoning and compositional question answering},
  author={Hudson, Drew A and Manning, Christopher D},
  booktitle={CVPR},
  year={2019}
}

@inproceedings{yue2024mmmu,
  title={Mmmu: A massive multi-discipline multimodal understanding and reasoning benchmark for expert agi},
  author={Yue, Xiang and Ni, Yuansheng and Zhang, Kai and Zheng, Tianyu and Liu, Ruoqi and Zhang, Ge and Stevens, Samuel and Jiang, Dongfu and Ren, Weiming and Sun, Yuxuan and others},
  booktitle={CVPR},
  year={2024}
}

@inproceedings{liu2024mmbench,
  title={Mmbench: Is your multi-modal model an all-around player?},
  author={Liu, Yuan and Duan, Haodong and Zhang, Yuanhan and Li, Bo and Zhang, Songyang and Zhao, Wangbo and Yuan, Yike and Wang, Jiaqi and He, Conghui and Liu, Ziwei and others},
  booktitle={ECCV},
  year={2024}
}

@inproceedings{fu2026mme,
  title={Mme: A comprehensive evaluation benchmark for multimodal large language models},
  author={Fu, Chaoyou and Chen, Peixian and Shen, Yunhang and Qin, Yulei and Zhang, Mengdan and Lin, Xu and Yang, Jinrui and Zheng, Xiawu and Li, Ke and Sun, Xing and others},
  booktitle={NeurIPS Track on Datasets and Benchmarks},
  year={2025}
}

@inproceedings{li2023evaluating,
  title={Evaluating object hallucination in large vision-language models},
  author={Li, Yifan and Du, Yifan and Zhou, Kun and Wang, Jinpeng and Zhao, Xin and Wen, Ji-Rong},
  booktitle={EMNLP},
  year={2023}
}

@inproceedings{singh2019towards,
  title={Towards vqa models that can read},
  author={Singh, Amanpreet and Natarajan, Vivek and Shah, Meet and Jiang, Yu and Chen, Xinlei and Batra, Dhruv and Parikh, Devi and Rohrbach, Marcus},
  booktitle={CVPR},
  year={2019}
}

@article{liu2024ocrbench,
  title={Ocrbench: on the hidden mystery of ocr in large multimodal models},
  author={Liu, Yuliang and Li, Zhang and Huang, Mingxin and Yang, Biao and Yu, Wenwen and Li, Chunyuan and Yin, Xu-Cheng and Liu, Cheng-Lin and Jin, Lianwen and Bai, Xiang},
  journal={Science China Information Sciences},
  year={2024}
}

@inproceedings{kim2026zoo,
  title={Zoo-prune: training-free token pruning via zeroth-order gradient estimation in vision-language models},
  author={Kim, Youngeun and Zhang, Youjia and Liu, Huiling and Jung, Aecheon and Lee, Sunwoo and Hong, Sungeun},
  booktitle={CVPR},
  year={2026}
}

@inproceedings{bengio2009curriculum,
  title={Curriculum learning},
  author={Bengio, Yoshua and Louradour, J{\'e}r{\^o}me and Collobert, Ronan and Weston, Jason},
  booktitle={ICML},
  year={2009}
}

@inproceedings{li2024mvbench,
  title={Mvbench: A comprehensive multi-modal video understanding benchmark},
  author={Li, Kunchang and Wang, Yali and He, Yinan and Li, Yizhuo and Wang, Yi and Liu, Yi and Wang, Zun and Xu, Jilan and Chen, Guo and Luo, Ping and others},
  booktitle={CVPR},
  year={2024}
}

@inproceedings{wang2025muirbench,
  title={Muirbench: A comprehensive benchmark for robust multi-image understanding},
  author={Wang, Fei and Fu, Xingyu and Huang, James Y and Li, Zekun and Liu, Qin and Liu, Xiaogeng and Ma, Mingyu Derek and Xu, Nan and Zhou, Wenxuan and Zhang, Kai and others},
  booktitle={ICLR},
  year={2025}
}

@inproceedings{zhang2026deep,
  title={Deep video discovery: Agentic search with tool use for long-form video understanding},
  author={Zhang, Xiaoyi and Jia, Zhaoyang and Guo, Zongyu and Li, Jiahao and Li, Bin and Li, Houqiang and Lu, Yan},
  booktitle={NeurIPS},
  year={2025}
}

@article{hurst2024gpt,
  title={Gpt-4o system card},
  author={Hurst, Aaron and Lerer, Adam and Goucher, Adam P and Perelman, Adam and Ramesh, Aditya and Clark, Aidan and Ostrow, AJ and Welihinda, Akila and Hayes, Alan and Radford, Alec and others},
  journal={arXiv preprint arXiv:2410.21276},
  year={2024}
}

@article{bai2025qwen3,
  title={Qwen3-vl technical report},
  author={Bai, Shuai and Cai, Yuxuan and Chen, Ruizhe and Chen, Keqin and Chen, Xionghui and Cheng, Zesen and Deng, Lianghao and Ding, Wei and Gao, Chang and Ge, Chunjiang and others},
  journal={arXiv preprint arXiv:2511.21631},
  year={2025}
}

@article{team2025kimi,
  title={Kimi-vl technical report},
  author={Team, Kimi and Du, Angang and Yin, Bohong and Xing, Bowei and Qu, Bowen and Wang, Bowen and Chen, Cheng and Zhang, Chenlin and Du, Chenzhuang and Wei, Chu and others},
  journal={arXiv preprint arXiv:2504.07491},
  year={2025}
}

@article{wang2025internvl3,
  title={Internvl3. 5: Advancing open-source multimodal models in versatility, reasoning, and efficiency},
  author={Wang, Weiyun and Gao, Zhangwei and Gu, Lixin and Pu, Hengjun and Cui, Long and Wei, Xingguang and Liu, Zhaoyang and Jing, Linglin and Ye, Shenglong and Shao, Jie and others},
  journal={arXiv preprint arXiv:2508.18265},
  year={2025}
}

@inproceedings{hu2024matryoshka,
  title={Matryoshka query transformer for large vision-language models},
  author={Hu, Wenbo and Dou, Zi-Yi and Li, Liunian H and Kamath, Amita and Peng, Nanyun and Chang, Kai-Wei},
  booktitle={NeurIPS},
  year={2024}
}

@article{menendez1997jensen,
  title={The jensen-shannon divergence},
  author={Men{\'e}ndez, Mar{\'\i}a Luisa and Pardo, Julio Angel and Pardo, Leandro and Pardo, Mar{\'\i}a del C},
  journal={Journal of the Franklin Institute},
  year={1997}
}

@inproceedings{feng2026onethinker,
  title={Onethinker: All-in-one reasoning model for image and video},
  author={Feng, Kaituo and Zhang, Manyuan and Li, Hongyu and Fan, Kaixuan and Chen, Shuang and Jiang, Yilei and Zheng, Dian and Sun, Peiwen and Zhang, Yiyuan and Sun, Haoze and others},
  booktitle={CVPR},
  year={2026}
}

@inproceedings{deitke2025molmo,
  title={Molmo and pixmo: Open weights and open data for state-of-the-art vision-language models},
  author={Deitke, Matt and Clark, Christopher and Lee, Sangho and Tripathi, Rohun and Yang, Yue and Park, Jae Sung and Salehi, Mohammadreza and Muennighoff, Niklas and Lo, Kyle and Soldaini, Luca and others},
  booktitle={CVPR},
  year={2025}
}

@inproceedings{liu2023visual,
  title={Visual instruction tuning},
  author={Liu, Haotian and Li, Chunyuan and Wu, Qingyang and Lee, Yong Jae},
  booktitle={NeurIPS},
  year={2023}
}

@inproceedings{li2025chemvlm,
  title={Chemvlm: Exploring the power of multimodal large language models in chemistry area},
  author={Li, Junxian and Zhang, Di and Wang, Xunzhi and Hao, Zeying and Lei, Jingdi and Tan, Qian and Zhou, Cai and Liu, Wei and Yang, Yaotian and Xiong, Xinrui and others},
  booktitle={Proceedings of the AAAI Conference on Artificial Intelligence},
  volume={39},
  pages={415--423},
  year={2025}
}

@article{wei2026zooming,
  title={Zooming without Zooming: Region-to-Image Distillation for Fine-Grained Multimodal Perception},
  author={Wei, Lai and He, Liangbo and Lan, Jun and Dong, Lingzhong and Cai, Yutong and Li, Siyuan and Zhu, Huijia and Wang, Weiqiang and Kong, Linghe and Wang, Yue and others},
  journal={arXiv preprint arXiv:2602.11858},
  year={2026}
}
\bibliographystyle{iclr2027_conference}

\appendix

\clearpage
\raggedbottom
\section{Training Data Details}
\label{app:data}

\paragraph{Training mixture.}
The distribution of LT-14K are in Table~\ref{tab:app_data}. 
A deterministic balanced sampler realizes these quotas over 175 batches of 80 questions. The Vision-OPD source uses sampling seed 20260830; the remaining sources use 20260824. We also carefully check the training data, and confirm that no data leakage happens.

\begin{table}[H]
\centering\small
\caption{Composition of the 14,000-question training set.}
\label{tab:app_data}
\begin{tabular}{@{}lr@{}}
\toprule
Source / task bucket & Questions \\
\midrule
OneThinker: multiple choice & 3,000 \\
OneThinker: mathematics & 1,500 \\
OneThinker: numerical answer & 900 \\
OneThinker: OCR & 900 \\
OneThinker: regression & 900 \\
PixMo Ask Model Anything & 3,000 \\
LLaVA v1.5 mixture & 2,800 \\
Vision-OPD full-image multiple choice & 1,000 \\
\midrule
Total & 14,000 \\
\bottomrule
\end{tabular}
\end{table}

\section{Implementation and Training Details}
\label{app:implementation}

\paragraph{Student and teacher.}
We initialize the student and teacher from Qwen3.5-4B. The teacher is a frozen copy of the initial model and runs in evaluation mode without gradients. Both receive the same clean full image and question, with the teacher using the dense visual representation. We train the student language model, vision encoder, and native visual merger with full-parameter updates.

\paragraph{Visual token selection.}
We adapt CDPruner to Qwen3.5 and apply it after the native visual merger and before the language model. The parameter-free selector combines question-conditioned relevance with visual diversity through a matrix-free greedy conditional determinantal-point-process procedure in FP32. Selection runs without gradients, breaks ties by original token index, and preserves the selected embeddings' original multimodal rotary positions.

\paragraph{Distillation objective.}
Training uses student-generated trajectories and distributional on-policy distillation. We use a Jensen-Shannon objective with mixture coefficient 0.5. At each response position, the support is the union of the student's top 100 tokens, the teacher's top 100 tokens, and the sampled token, with the remaining probability mass represented by one aggregated tail category, following~\citep{yuan2026vision}. Loss reduction averages valid response-token losses within each sequence and then averages over sequences.

\begin{table}[H]
\centering\small
\caption{Training configuration of the 14K curriculum model. Learning rates are the values after warmup.}
\label{tab:app_train}
\begin{tabular}{@{}p{0.48\linewidth}p{0.46\linewidth}@{}}
\toprule
Item & Value \\
\midrule
Training questions / epochs / optimizer updates & 14,000 / 1 / 175 \\
Questions per update / rollouts per question & 80 / 8 \\
Optimizer & AdamW \\
Language model / native merger learning rate & $1\times10^{-6}$ / $1\times10^{-6}$ \\
Vision encoder learning rate & $2\times10^{-7}$ \\
Adam betas / epsilon & $(0.9,0.999)$ / $10^{-8}$ \\
Weight decay / gradient norm clip & 0.01 / 1.0 \\
Learning-rate schedule & 14-update warmup, then constant \\
Initialization seed & 42 \\
Compute / master parameter precision & BF16 / FP32 \\
Parallelism & FSDP; 8 data-parallel ranks \\
Training hardware & 8 NVIDIA A100 80GB PCIe GPUs \\
Gradient checkpointing & Non-reentrant, enabled for student \\
Rollout backend & Hugging Face \\
Training sampling & Temperature 1; top-$p$ 1; no top-$k$ truncation \\
Maximum response length & 1,024 new tokens \\
Maximum prompt / model length & 24,576 / 26,624 tokens \\
Image pixel limits & 65,536 to 16,777,216 \\
\bottomrule
\end{tabular}
\end{table}

\paragraph{Execution details.}
Training comprises 175 updates and 112,000 generated trajectories; evaluation uses the final checkpoint at update 175. The image processor applies the model's native resize before patchification, with patch size 16 and native spatial merge size 2. Bias, normalization, and parameter tensors with fewer than two dimensions receive zero weight decay.

Each student update microbatch contains up to seven trajectories, and each teacher microbatch up to two. The Hugging Face rollout model is replicated, with a decode-batch cap of 80 and CPU offload between phases. We use Python 3.12.13, PyTorch 2.10.0 (CUDA 12.6), Transformers 5.5.0, FlashAttention 2.8.3, and Ray 2.53.0.

\section{Training Prompts}
\label{app:training_prompts}

We append the task-specific response instruction below to the original question. Braced fields denote runtime substitutions, and the image placeholder denotes a multimodal image input. The chat template fixes the assistant prefix to an empty thinking block followed by an open answer tag:

\begin{insbox}
\small
\begin{Verbatim}
<|im_start|>assistant
<think>

</think>

<answer>
\end{Verbatim}
\end{insbox}

Generation stops at the first exact closing tag \texttt{</answer>}, with an EOS fallback and the configured length cap. The opening tag belongs to prompt context; the closing tag is sampled, and tokens after stopping do not contribute to the loss.

\par\addvspace{9pt}\noindent\begin{minipage}{\linewidth}
\noindent\textbf{Task prompt template.}\par\nobreak\vspace{5pt}
The original question and options are preserved. The task instruction is substituted verbatim from the following list.
\begin{insbox}
\small
\begin{Verbatim}
<image>
{original_question_and_options}

Response requirements: {task_instruction}
\end{Verbatim}
\end{insbox}
\par\vspace{4pt}
\end{minipage}\par
\par\addvspace{9pt}\noindent\begin{minipage}{\linewidth}
\noindent\textbf{T1. Multiple choice.}\par\nobreak\vspace{5pt}
\begin{insbox}
\small
\begin{Verbatim}[fontsize=\small,breaklines=true,breakanywhere=true,breaksymbolleft={},breaksymbolright={},frame=none,framesep=0pt,framerule=0pt,baselinestretch=1.08,rulecolor=\color{black!55},listparameters={\setlength{\topsep}{0pt}\setlength{\partopsep}{0pt}}]
Output protocol: Preserve every content and option-label requirement in the original question. The assistant response is already prefixed with the exact opening <answer> delimiter. Do not emit another <answer>. Decide silently, continue with exactly one option label, then emit exactly one </answer> and stop immediately. Do not output reasoning, commentary, multiple labels, repetition, or revisions.
\end{Verbatim}
\end{insbox}
\par\vspace{4pt}
\end{minipage}\par
\par\addvspace{9pt}\noindent\begin{minipage}{\linewidth}
\noindent\textbf{T2. Mathematics.}\par\nobreak\vspace{5pt}
\begin{insbox}
\small
\begin{Verbatim}[fontsize=\small,breaklines=true,breakanywhere=true,breaksymbolleft={},breaksymbolright={},frame=none,framesep=0pt,framerule=0pt,baselinestretch=1.08,rulecolor=\color{black!55},listparameters={\setlength{\topsep}{0pt}\setlength{\partopsep}{0pt}}]
Output protocol: Preserve every content and internal-format requirement in the original question. The assistant response is already prefixed with the exact opening <answer> delimiter. Do not emit another <answer>. Unless the original question explicitly requests reasoning or derivation, solve silently and continue with only the final response. If it explicitly requests reasoning or derivation, use at most four short steps. Then emit exactly one </answer> and stop immediately; do not explore alternatives, repeat, or revise.
\end{Verbatim}
\end{insbox}
\par\vspace{4pt}
\end{minipage}\par
\par\addvspace{9pt}\noindent\begin{minipage}{\linewidth}
\noindent\textbf{T3. Numerical answers.}\par\nobreak\vspace{5pt}
\begin{insbox}
\small
\begin{Verbatim}[fontsize=\small,breaklines=true,breakanywhere=true,breaksymbolleft={},breaksymbolright={},frame=none,framesep=0pt,framerule=0pt,baselinestretch=1.08,rulecolor=\color{black!55},listparameters={\setlength{\topsep}{0pt}\setlength{\partopsep}{0pt}}]
Output protocol: Preserve every precision and sign requirement in the original question. The assistant response is already prefixed with the exact opening <answer> delimiter. Do not emit another <answer>. Compute silently, continue with exactly one final scalar number without units, brackets, or thousands separators, then emit exactly one </answer> and stop immediately. Do not output reasoning, commentary, repetition, or revisions.
\end{Verbatim}
\end{insbox}
\par\vspace{4pt}
\end{minipage}\par
\par\addvspace{9pt}\noindent\begin{minipage}{\linewidth}
\noindent\textbf{T4. OCR and transcription.}\par\nobreak\vspace{5pt}
\begin{insbox}
\begin{Verbatim}[fontsize=\small,breaklines=true,breakanywhere=true,breaksymbolleft={},breaksymbolright={},frame=none,framesep=0pt,framerule=0pt,baselinestretch=1.08,rulecolor=\color{black!55},listparameters={\setlength{\topsep}{0pt}\setlength{\partopsep}{0pt}}]
Output protocol: Preserve every content and internal-format requirement in the original question, including requests for transcription, caption, description, multiple lines, or LaTeX. The assistant response is already prefixed with the exact opening <answer> delimiter. Do not emit another <answer>. Continue with the complete requested response, then emit exactly one </answer> and stop immediately. Do not add unrelated commentary, repeat, or revise.
\end{Verbatim}
\end{insbox}
\par\vspace{4pt}
\end{minipage}\par
\par\addvspace{9pt}\noindent\begin{minipage}{\linewidth}
\noindent\textbf{T5. Regression.}\par\nobreak\vspace{5pt}
\begin{insbox}
\begin{Verbatim}[fontsize=\small,breaklines=true,breakanywhere=true,breaksymbolleft={},breaksymbolright={},frame=none,framesep=0pt,framerule=0pt,baselinestretch=1.08,rulecolor=\color{black!55},listparameters={\setlength{\topsep}{0pt}\setlength{\partopsep}{0pt}}]
Output protocol: Preserve every precision and sign requirement in the original question. The assistant response is already prefixed with the exact opening <answer> delimiter. Do not emit another <answer>. Estimate silently, continue with exactly one final scalar number without units, brackets, or thousands separators, then emit exactly one </answer> and stop immediately. Do not output reasoning, commentary, repetition, or revisions.
\end{Verbatim}
\end{insbox}
\par\vspace{4pt}
\end{minipage}\par
\par\addvspace{9pt}\noindent\begin{minipage}{\linewidth}
\noindent\textbf{T6. Open-ended VQA.}\par\nobreak\vspace{5pt}
\begin{insbox}
\small
\begin{Verbatim}[fontsize=\small,breaklines=true,breakanywhere=true,breaksymbolleft={},breaksymbolright={},frame=none,framesep=0pt,framerule=0pt,baselinestretch=1.08,rulecolor=\color{black!55},listparameters={\setlength{\topsep}{0pt}\setlength{\partopsep}{0pt}}]
Output protocol: Preserve every content and internal-format requirement in the original question. The assistant response is already prefixed with the exact opening <answer> delimiter. Do not emit another <answer>. Continue with a concise but complete requested response, then emit exactly one </answer> and stop immediately. Do not add unrelated commentary, repeat, or revise.
\end{Verbatim}
\par\vspace{4pt}
\end{insbox}
\end{minipage}\par
\par\addvspace{9pt}\noindent\begin{minipage}{\linewidth}
\noindent\textbf{T7. TextVQA.}\par\nobreak\vspace{5pt}
\begin{insbox}
\small
\begin{Verbatim}[fontsize=\small,breaklines=true,breakanywhere=true,breaksymbolleft={},breaksymbolright={},frame=none,framesep=0pt,framerule=0pt,baselinestretch=1.08,rulecolor=\color{black!55},listparameters={\setlength{\topsep}{0pt}\setlength{\partopsep}{0pt}}]
Output protocol: Read the visible text carefully and preserve the answer wording needed by the question. The assistant response is already prefixed with the exact opening <answer> delimiter. Do not emit another <answer>. Continue with the shortest complete answer supported by the image, then emit exactly one </answer> and stop immediately. Do not add explanation unless the original question explicitly requests it; do not repeat or revise.
\end{Verbatim}
\end{insbox}
\par\vspace{4pt}
\end{minipage}\par
\par\addvspace{9pt}\noindent\begin{minipage}{\linewidth}
\noindent\textbf{T8. Full-image multiple choice.}\par\nobreak\vspace{5pt}
\begin{insbox}
\small
\begin{Verbatim}[fontsize=\small,breaklines=true,breakanywhere=true,breaksymbolleft={},breaksymbolright={},frame=none,framesep=0pt,framerule=0pt,baselinestretch=1.08,rulecolor=\color{black!55},listparameters={\setlength{\topsep}{0pt}\setlength{\partopsep}{0pt}}]
<image>
Inspect the full image carefully. Use all image regions relevant to the question, including fine-grained and peripheral details.
Question: {question}
Choices:
A. {option_A}
B. {option_B}
C. {option_C}
D. {option_D}

Response requirements: The assistant response is already prefixed with the exact opening <answer> delimiter. Decide silently, continue with exactly one capital option letter (A, B, C, or D), then emit exactly one </answer> and stop immediately. Do not emit reasoning, commentary, another <answer>, or any other text.
\end{Verbatim}
\end{insbox}
\par\vspace{4pt}
\end{minipage}\par

\section{Evaluation Protocol and Prompt Templates}
\label{app:evaluation}

\subsection{Decoding and Scoring}
\paragraph{Input and decoding.}
Evaluation uses the final compressed checkpoint with its exported processor and chat template. The system message is \texttt{You are a helpful assistant.} Image inputs precede the dataset-specific text, followed by the same empty-thinking and open-answer assistant prefix used in training. Decoding is greedy, with one response per input, stopping at the first closing-answer tag or the generation cap in Table~\ref{tab:app_eval}. Questions with multiple images preserve all required image inputs; compression applies to the image spans.

\textbf{Evaluation implementation.} We use Lmms-eval~\citep{zhang2025lmms} and VLMEvalKit~\citep{duan2024vlmevalkit} for the chosen benchmarks. V$^*$Bench reports option accuracy. HR-Bench reports a macro average over cycle categories. MMMU reports instance accuracy. MMBench measures circular consistency across option permutations of 1,164 base-question groups. GQA uses normalized exact match, and TextVQA uses EvalAI-style VQA soft accuracy. OCRBench reports a total out of 1,000. POPE reports macro accuracy across adversarial, popular, and random subsets. MME sums per-category scores and paired-question scores, with perception and cognition subtotals.

\begin{table}[H]
\centering
\caption{Evaluated splits, row counts, and maximum new tokens. All tasks also use the shared closing-answer-tag stop.}
\label{tab:app_eval}
\begin{tabular}{@{}p{0.26\linewidth}p{0.32\linewidth}rr@{}}
\toprule
Benchmark & Split / setting & Rows & New tokens \\
\midrule
V$^*$Bench & test & 191 & 16 \\
HR-Bench 4K & 4K & 800 & 1024 \\
GQA & Testdev balanced & 12,578 & 16 \\
MMMU & validation & 900 & 128 \\
MMBench EN Circular & EN dev (circular) & 4,329 & 1024 \\
POPE & Three COCO subsets & 9,000 & 128 \\
TextVQA & Validation 0.5.1 & 5,000 & 1024 \\
OCRBench & test & 1,000 & 128 \\
MME & test & 2,374 & 16 \\
\bottomrule
\end{tabular}
\end{table}

\subsection{Evaluation Prompt Templates}
The following templates show the question formats and answer instructions; repeated instructions are consolidated for presentation. Option-list fields contain the original ordered choices in \texttt{A. value}, \texttt{B. value}, and subsequent lines.

\par\addvspace{9pt}\noindent\begin{minipage}{\linewidth}
\noindent\textbf{E1. V* Bench.}\par\nobreak\vspace{5pt}
\begin{insbox}
\begin{Verbatim}
{question}
{lettered_options}
Answer with the option's letter from the given choices directly.
\end{Verbatim}
\end{insbox}
\par\vspace{4pt}
\end{minipage}\par

\par\addvspace{9pt}\noindent\begin{minipage}{\linewidth}
\noindent\textbf{E2. HR-Bench 4K and 8K.}\par\nobreak\vspace{5pt}
\begin{insbox}
\begin{Verbatim}[fontsize=\small,breaklines=true,breakanywhere=true,breaksymbolleft={},breaksymbolright={},frame=none,framesep=0pt,framerule=0pt,baselinestretch=1.08,rulecolor=\color{black!55},listparameters={\setlength{\topsep}{0pt}\setlength{\partopsep}{0pt}}]
{question.strip()}
{lettered_options_with_trailing_newline}Answer the option letter directly.
\end{Verbatim}
\end{insbox}
\par\vspace{4pt}
\end{minipage}\par

\par\addvspace{9pt}\noindent\begin{minipage}{\linewidth}
\noindent\textbf{E3. MMMU, multiple choice.}\par\nobreak\vspace{5pt}
\begin{insbox}
\small
\begin{Verbatim}[fontsize=\small,breaklines=true,breakanywhere=true,breaksymbolleft={},breaksymbolright={},frame=none,framesep=0pt,framerule=0pt,baselinestretch=1.08,rulecolor=\color{black!55},listparameters={\setlength{\topsep}{0pt}\setlength{\partopsep}{0pt}}]
Question: {question}
Options:
{lettered_options}
Answer with the option letter only.
\end{Verbatim}
\end{insbox}
\par\vspace{4pt}
\end{minipage}\par

\par\addvspace{9pt}\noindent\begin{minipage}{\linewidth}
\noindent\textbf{E4. MMMU, open-ended.}\par\nobreak\vspace{5pt}
\begin{insbox}
\small
\begin{Verbatim}[fontsize=\small,breaklines=true,breakanywhere=true,breaksymbolleft={},breaksymbolright={},frame=none,framesep=0pt,framerule=0pt,baselinestretch=1.08,rulecolor=\color{black!55},listparameters={\setlength{\topsep}{0pt}\setlength{\partopsep}{0pt}}]
Question: {question}
Options:
{lettered_options}
Please answer the question directly.
\end{Verbatim}
\end{insbox}
\par\vspace{4pt}
This template covers 53 open-ended questions, retaining the Options header even for an empty list; text-only examples remain text-only.
\end{minipage}\par

\par\addvspace{9pt}\noindent\begin{minipage}{\linewidth}
\noindent\textbf{E5. OCRBench.}\par\nobreak\vspace{5pt}
\begin{insbox}
\small
\begin{Verbatim}
{question.strip()}
\end{Verbatim}
\end{insbox}
\par\vspace{4pt}
No additional user instruction or response suffix is added.
\end{minipage}\par

\par\addvspace{9pt}\noindent\begin{minipage}{\linewidth}
\noindent\textbf{E6. MMBench EN Circular.}\par\nobreak\vspace{5pt}
\begin{insbox}
\small
\begin{Verbatim}[fontsize=\small,breaklines=true,breakanywhere=true,breaksymbolleft={},breaksymbolright={},frame=none,framesep=0pt,framerule=0pt,baselinestretch=1.08,rulecolor=\color{black!55},listparameters={\setlength{\topsep}{0pt}\setlength{\partopsep}{0pt}}]
{hint_and_space_if_present}{question} There are several options:
{lettered_options}

Answer with the option's letter from the given choices directly.
\end{Verbatim}
\end{insbox}
\par\vspace{4pt}
The template includes the optional hint and valid A--E dataset options.
\end{minipage}\par

\par\addvspace{9pt}\noindent\begin{minipage}{\linewidth}
\noindent\textbf{E7. TextVQA.}\par\nobreak\vspace{5pt}
\begin{insbox}
\small
\begin{Verbatim}[fontsize=\small,breaklines=true,breakanywhere=true,breaksymbolleft={},breaksymbolright={},frame=none,framesep=0pt,framerule=0pt,baselinestretch=1.08,rulecolor=\color{black!55},listparameters={\setlength{\topsep}{0pt}\setlength{\partopsep}{0pt}}]
{question.capitalize()}
Answer the question using a single word or phrase.
\end{Verbatim}
\end{insbox}
\par\vspace{4pt}
\end{minipage}\par

\par\addvspace{9pt}\noindent\begin{minipage}{\linewidth}
\noindent\textbf{E8. POPE and GQA.}\par\nobreak\vspace{5pt}
\begin{insbox}
\small
\begin{Verbatim}[fontsize=\small,breaklines=true,breakanywhere=true,breaksymbolleft={},breaksymbolright={},frame=none,framesep=0pt,framerule=0pt,baselinestretch=1.08,rulecolor=\color{black!55},listparameters={\setlength{\topsep}{0pt}\setlength{\partopsep}{0pt}}]
{question.strip()}
Answer the question using a single word or phrase.
\end{Verbatim}
\end{insbox}
\par\vspace{4pt}
\end{minipage}\par

\par\addvspace{9pt}\noindent\begin{minipage}{\linewidth}
\noindent\textbf{E9. MME.}\par\nobreak\vspace{5pt}
\begin{insbox}
\begin{Verbatim}[fontsize=\small,breaklines=true,breakanywhere=true,breaksymbolleft={},breaksymbolright={},frame=none,framesep=0pt,framerule=0pt,baselinestretch=1.08,rulecolor=\color{black!55},listparameters={\setlength{\topsep}{0pt}\setlength{\partopsep}{0pt}}]
{question.strip().replace(" Please answer yes or no.", "")}
Answer the question using a single word or phrase.
\end{Verbatim}
\end{insbox}
\par\vspace{4pt}
\end{minipage}\par

\section{Efficiency and Training Cost}
\label{app:additional_evaluation}

\subsection{Efficiency Measurement}
We measure efficiency on 305 TextVQA questions covering 294 images. The subset contains questions whose most frequent human answer has at least six Qwen-tokenizer tokens after lowercasing and whitespace normalization. Each question is measured five times per method using BF16, FlashAttention 2, batch size one, and a common CUDA-graph backend. Decoding replays a shared 128-token ground-truth-conditioned reference with 127 incremental forward passes. Compressed models use the 5\% retention rule. HiPrune prunes before the first language-model layer; the EPIC implementation prunes before the third layer and uses a trained adapter merged into the base model.

Request timing runs from prepared CPU tensors to decoded output text, covering transfer, vision encoding, token selection, language-model prefill, reference decoding, and synchronization. Measurements are averaged over repetitions within each question and then across questions. Table~\ref{tab:measurement_scope} defines the resource quantities. We use decimal units: one MB is $10^6$ bytes and one TFLOP is $10^{12}$ operations, with each multiply--accumulate counted as two operations.

\begin{table}[H]
\centering\small
\caption{Resource quantities used in the efficiency evaluation.}
\label{tab:measurement_scope}
\begin{tabular}{@{}p{0.25\linewidth}p{0.70\linewidth}@{}}
\toprule Quantity & Definition \\
\midrule
LLM prefill FLOPs & Analytical cost of language-model prefill and the final-token output head. \\
Whole-prefill FLOPs & Analytical cost of vision encoding, selection, and language-model prefill. \\
Full-attention KV & Key/value tensors in the eight full-attention layers after prefill, including text context. \\
Total decode state & Full-attention KV plus convolutional and recurrent states in the 24 linear-attention layers. \\
Replay request time & Elapsed time from prepared CPU tensors to decoded text under the fixed reference replay. \\
\bottomrule
\end{tabular}
\end{table}

\section{Explanations of Different training strategies}
\label{app:strategy}
We further examine different strategies for constructing the full-token
teacher during LT-OPD. In all variants, the student generates the on-policy
trajectory with compressed visual tokens, while the teacher receives the
full visual tokens and provides next-token distributional supervision on
the same student-generated prefixes.

\textbf{Initial-policy teacher.}
Our default strategy keeps the teacher fixed at the pretrained initialization
$\theta_0$ throughout training, i.e., $\pi_T=\pi_{\theta_0}$.
This provides a stationary supervision target and prevents the teacher from
co-adapting with the compressed student. This strategy is also consistent
with prior on-policy self-distillation practice~\citep{zhao2026self}.

\textbf{Trust-region regularization.}
We also consider an adaptive teacher regularized toward the initial policy.
At training step $s$, the teacher distribution is constructed by interpolating
the current full-token policy with the frozen reference policy in log-probability
space:
\begin{equation}
\pi_T^{\mathrm{TR}}(a\mid x,s)
\propto
\exp\!\left(
(1-\lambda)\log \pi_{\theta_0}(a\mid x,s)
+
\lambda\log \pi_{\theta_s}(a\mid x,s)
\right),
\end{equation}
where $\lambda\in[0,1]$ controls how strongly the teacher follows the
current policy. This allows the teacher to evolve during training while
constraining it from drifting excessively from the pretrained model.

\textbf{EMA teacher.}
Alternatively, we maintain a separate teacher whose parameters are updated
as an exponential moving average (EMA) of the student:
\begin{equation}
\theta_T^{(s)}
=
(1-\alpha)\theta_T^{(s-1)}
+
\alpha\theta_s,
\end{equation}
with $\theta_T^{(0)}=\theta_0$. The EMA teacher changes more smoothly than
the current student and therefore provides a slowly evolving supervision
target. Similar EMA regularization has been adopted in recent multimodal
on-policy self-distillation~\citep{yuan2026vision}.

\par\addvspace{1.3ex}\noindent\begin{minipage}{\linewidth}
\section{Ablation on Training Objectives}
\label{app:objectives}
Table~\ref{tab:ablation1} reports ablations results for forward KL, reverse KL, and JSD. Notably, forward KL and reverse KL are two different training objectives described in~\cite{yuan2026vision}. Here we also list the detailed training objectives of them:

$\mathrm{forward \ KL}: \mathcal{L}_{\mathrm{FKL}}=\mathbb{E}_{s\sim d_{\pi_{\mathrm{S}}}}\left[D_{\mathrm{KL}}\!\left(\pi_{\mathrm{T}}(\cdot\mid s)\,\|\,\pi_{\mathrm{S}}(\cdot\mid s)\right)\right]=\mathbb{E}_{s\sim d_{\pi_{\mathrm{S}}}}\left[\sum_{a}\pi_{\mathrm{T}}(a\mid s)\log\frac{\pi_{\mathrm{T}}(a\mid s)}{\pi_{\mathrm{S}}(a\mid s)}\right.$,

$\mathrm{Reverse \ KL}: \mathcal{L}_{\mathrm{RKL}}=\mathbb{E}_{s\sim d_{\pi_{\mathrm{S}}}}\left[D_{\mathrm{KL}}\!\left(\pi_{\mathrm{S}}(\cdot\mid s)\,\|\,\pi_{\mathrm{T}}(\cdot\mid s)\right)\right]=\mathbb{E}_{s\sim d_{\pi_{\mathrm{S}}}}\left[\sum_{a}\pi_{\mathrm{S}}(a\mid s)\log\frac{\pi_{\mathrm{S}}(a\mid s)}{\pi_{\mathrm{T}}(a\mid s)}\right].
$

$\mathrm{JSD}: \mathcal{L}_{\mathrm{JSD}}=\mathbb{E}_{s\sim d_{\pi_{\mathrm{S}}}}\left[\frac{1}{2}D_{\mathrm{KL}}\!\left(\pi_{\mathrm{T}}(\cdot\mid s)\,\|\,m(\cdot\mid s)\right)+\frac{1}{2}D_{\mathrm{KL}}\!\left(\pi_{\mathrm{S}}(\cdot\mid s)\,\|\,m(\cdot\mid s)\right)\right],m(a\mid s)=\frac{\pi_{\mathrm{T}}(a\mid s)+\pi_{\mathrm{S}}(a\mid s)}{2}.$

\begin{table}[H]
\centering
\caption{Training-objective results. Full-token and JSD results are retained from the main text. Avg. is the same as Table~\ref{tab:baseline} in main text.}
\label{tab:ablation1}
\resizebox{\textwidth}{!}{%
\renewcommand{\arraystretch}{1.15}
\footnotesize\setlength{\tabcolsep}{2.2pt}\begin{tabular}{lcccccccccc}
\toprule
\textbf{Method}
& \textbf{V$^*$Bench}
& \textbf{HR-4K}
& \textbf{GQA}
& \textbf{MMMU}
& \textbf{MMB}
& \textbf{MME}
& \textbf{POPE}
& \textbf{TextVQA}
& \textbf{OCRB}
& \textbf{Avg.} \\
\midrule


\rowcolor{red!10} Qwen3.5-4B (\textit{Upper Bound})
        & 84.3 & 84.4 & 60.7  & 77.6 &  86.4
        & 2,300.8 & 89.3 & 81.2 & 850 & 100\% \\

Forward KL ($\mathrm{KL}(p_T \Vert p_S)$)
            & 70.2 & 68.8 & 59.5 & \textbf{52.2} & \textbf{82.2} &  1,568.2&  \textbf{89.1} &  62.5& \textbf{599} & 82.3\%\\
Reverse KL ($\mathrm{KL}(p_S \Vert p_T)$)
            & 70.2 & 71.3 &  \textbf{59.6} &  51.4&  82.0&  1,600.4&  \textbf{89.1} &  \textbf{63.6} & \textbf{599} & \textbf{82.8\%}\\
\rowcolor{orange!10}
JSD ($\beta = 0.5$)
& \textbf{74.9} & \textbf{73.0} & 54.8
& 50.0 & 78.9
& \textbf{2,077.0} & 86.3
& 56.0 & 535
& 82.3\% \\
\bottomrule
\end{tabular}%
}
\vspace{-3mm}
\end{table}
\end{minipage}\par

\section{Ablation on the Visual-Budget Curriculum}
\label{app:curriculum_ablation}
The visual-budget schedule keeps the retention at 25\% for the first 14 updates, decays it along a cosine curve, and then holds it at 5\% for updates 101--175. Table~\ref{tab:ablation_curriculum} isolates the two constant stages: the \emph{w/o final holding} variant keeps the warmup, the cosine decay, and the 5\% endpoint but removes the frozen 5\% hold, so the retention reaches 5\% only at the last update, while the \emph{w/o initial holding} variant removes the 25\% hold so that the decay starts immediately. Both variants otherwise follow the reference training configuration (same initialization seed, data order, and 175-update budget) and are evaluated with the protocol of Table~\ref{tab:app_eval} at 5\% retention.

\begin{table}[H]
\centering
\caption{Effect of removing the two constant stages of the visual-budget schedule. Row \emph{LT-OPD} is the full-schedule model from Table~\ref{tab:ablation1}. Avg. is the same as Table~\ref{tab:baseline} in main text.}
\label{tab:ablation_curriculum}
\resizebox{\textwidth}{!}{%
\renewcommand{\arraystretch}{1.15}
\footnotesize\setlength{\tabcolsep}{2.2pt}\begin{tabular}{lcccccccccc}
\toprule
\textbf{Method}
& \textbf{V$^*$Bench}
& \textbf{HR-4K}
& \textbf{GQA}
& \textbf{MMMU}
& \textbf{MMB}
& \textbf{MME}
& \textbf{POPE}
& \textbf{TextVQA}
& \textbf{OCRB}
& \textbf{Avg.} \\
\midrule
\rowcolor{orange!10}
LT-OPD
& \textbf{74.9} & \textbf{73.0} & \textbf{54.8} & \textbf{50.0} & \textbf{78.9}
& \textbf{2,077.0} & 86.3 & \textbf{56.0} & \textbf{535} & \textbf{82.3\%} \\
w/o final holding
& 72.3 & 69.1 & \textbf{54.8} & 49.8 & 77.7
& 2,025.0 & \textbf{86.4} & 55.2 & 524 & 80.7\% \\
w/o initial holding
& 68.1 & 69.7 & \textbf{54.8} & 49.7 & 78.2
& 2,049.3 & 86.3 & \textbf{56.0} & 531 & 80.6\% \\
\bottomrule
\end{tabular}%
}
\vspace{-3mm}
\end{table}

\section{Results on Multi-image Benchmarks}
\label{app:multi_image}
We test on two multi-image benchmarks, MuirBench and MVBench. Note that MuirBench is a multi-image-understanding benchmark, while MVBench is a video-understanding benchmark (16 frames uniform sampling). Table~\ref{tab:multi_image} reports the results. 

\begin{table}[H]
\centering
\caption{Performance on multi-image benchmarks. Two baselines keep 5\% visual tokens. Avg. is the same as Table~\ref{tab:baseline} in main text.}
\label{tab:multi_image}
\renewcommand{\arraystretch}{1.15}
\footnotesize\setlength{\tabcolsep}{2.2pt}\begin{tabular}{lccc}
\toprule
\textbf{Method}
& \textbf{MuirBench}
& \textbf{MVBench}
& \textbf{Avg.} \\
\midrule


\rowcolor{red!10} Qwen3.5-4B (\textit{Upper Bound})
        & 65.7 &  71.8 & 100\% \\

 CDPruner 
        & 39.1 & 51.6  & 65.7\% \\
 EPIC
        & 44.0 &  52.2 & 69.8\% \\
\rowcolor{orange!10} LT-OPD
        & \textbf{52.0} & \textbf{55.6}  & \textbf{78.3\%} \\
        
\bottomrule
\end{tabular}%

\end{table}

\par\addvspace{6pt}\noindent\begin{minipage}{\linewidth}
\section{Qualitative Case Studies}
\label{app:cases}
We compare HiPrune and EPIC outputs with the final LT-OPD checkpoint on three shared image-question items. We report evidence details in the image, and model predictions.

\subsection{Object Material in a Crowded Scene}
\begin{figure}[H]
\centering
\begin{minipage}[t]{0.74\linewidth}
\vspace{0pt}\includegraphics[width=\linewidth]{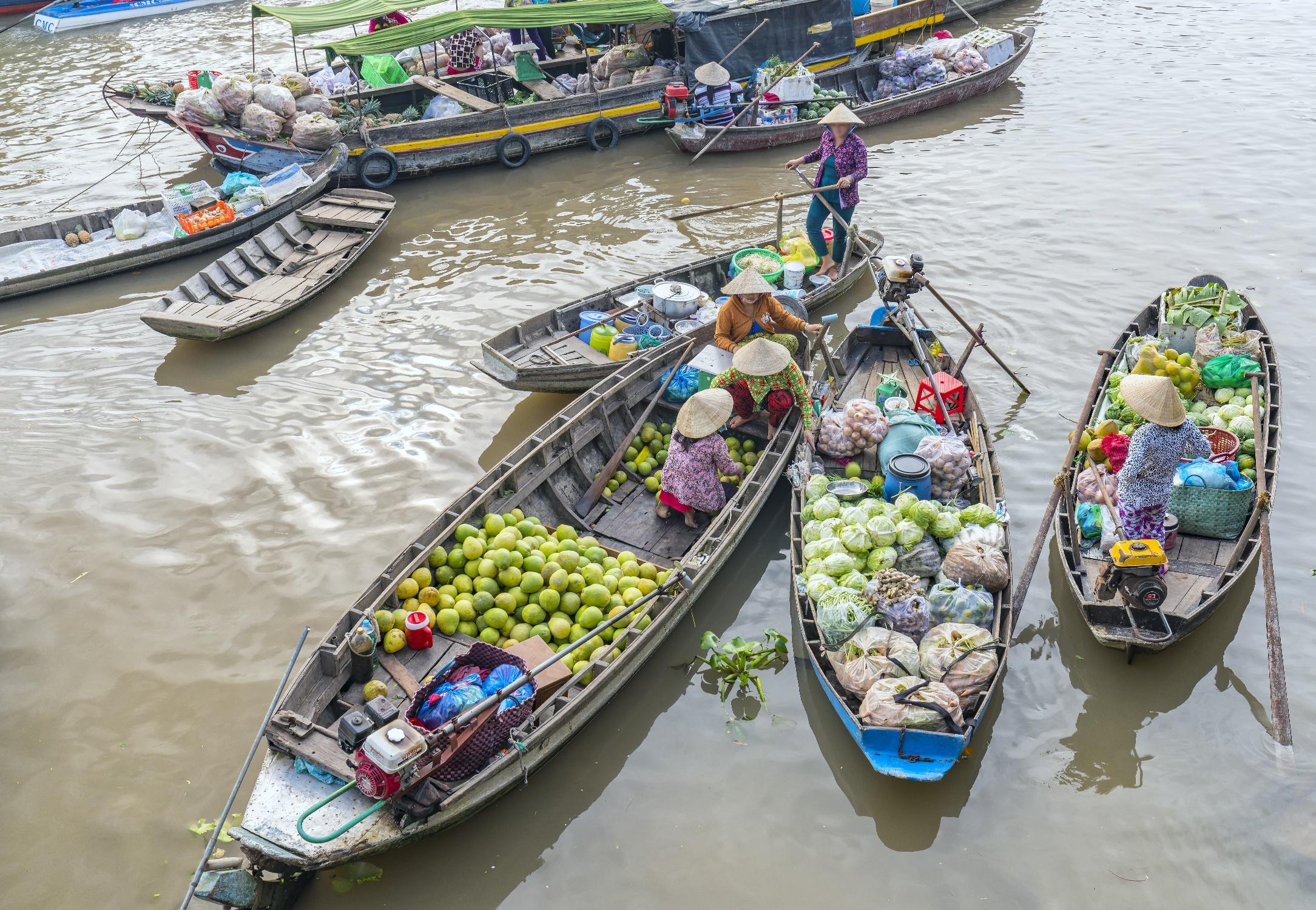}
\end{minipage}\hfill
\begin{minipage}[t]{0.23\linewidth}
\vspace{0pt}\small\textbf{Evidence detail}\par\smallskip
\includegraphics[width=\linewidth,trim={1480.0000bp 790.0000bp 549.0000bp 593.0000bp},clip]{images/reviewed_case_vstar_18.jpg}
\par\smallskip Reader-only enlargement of the red stool.
\end{minipage}
\par\medskip
\begin{minipage}{\linewidth}
\textbf{Question.} What is the material of the stool?\\[3pt]
A. bamboo \qquad B. steel \qquad C. plastic \qquad D. wood\\[5pt]
\textbf{Reference answer.} C (plastic)\par\medskip
\setlength{\tabcolsep}{8pt}\renewcommand{\arraystretch}{1.4}
\begin{tabular}{>{\raggedright\arraybackslash}p{0.18\linewidth}>{\raggedright\arraybackslash}p{0.57\linewidth}>{\raggedright\arraybackslash}p{\dimexpr0.25\linewidth-6\tabcolsep\relax}}
\toprule Method & Recorded answer text & Outcome \\
\midrule
\rowcolor{red!7}HiPrune & \texttt{A} & Incorrect \\
\rowcolor{red!7}EPIC & \texttt{D} & Incorrect \\
\rowcolor{green!7}LT-OPD & \texttt{C} & Correct \\
\bottomrule
\end{tabular}
\end{minipage}
\caption{V$^*$Bench item 18. A plastic stool appears among boats and produce. HiPrune and EPIC select bamboo and wood, respectively; LT-OPD correctly selects plastic.}
\label{fig:case_material}
\end{figure}
\end{minipage}\par

\clearpage
\subsection{A Small Attribute in a Large Scene}
\begin{figure}[H]
\centering
\begin{minipage}[t]{0.57\linewidth}
\vspace{0pt}\includegraphics[width=\linewidth]{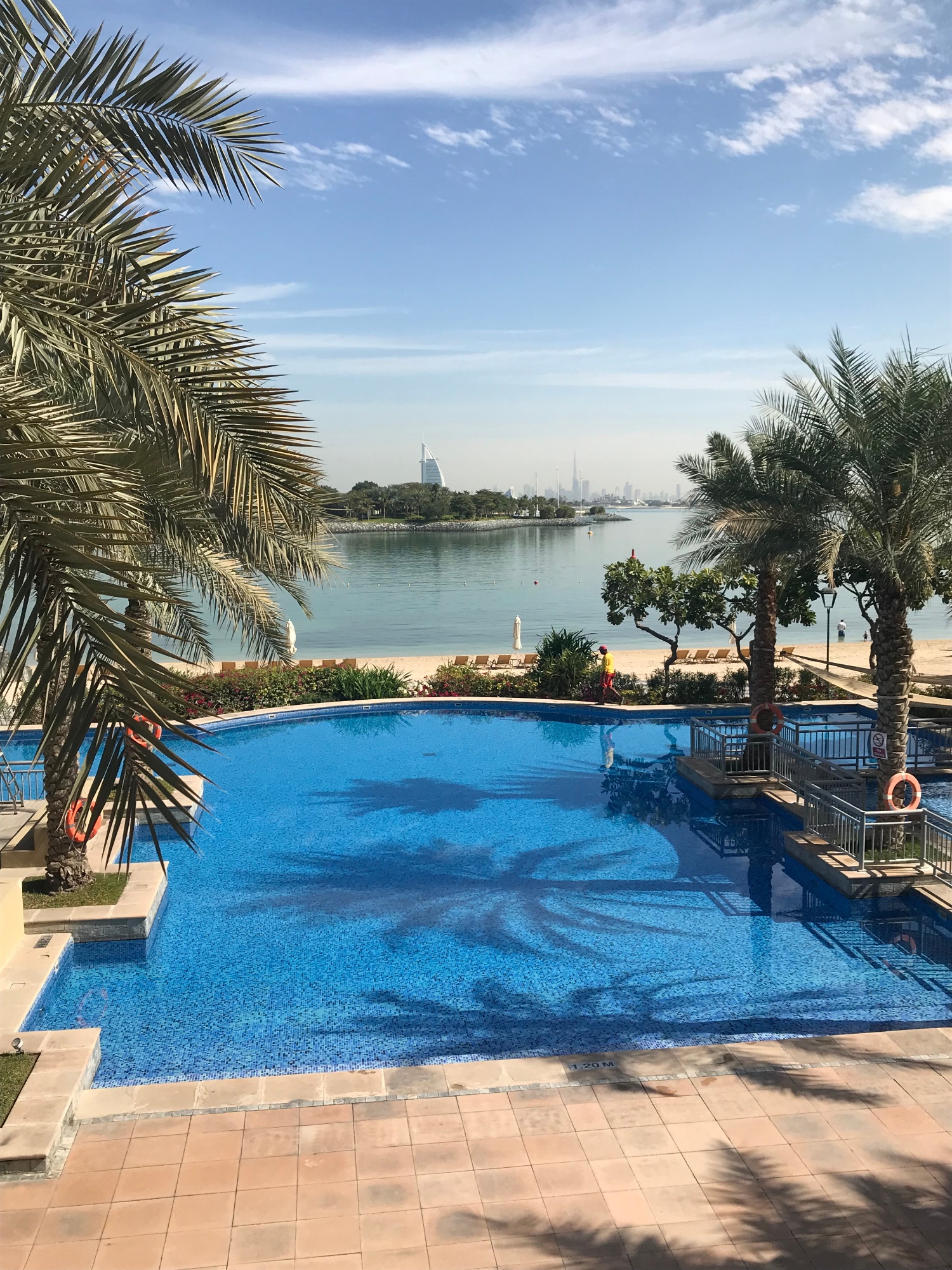}
\end{minipage}\hfill
\begin{minipage}[t]{0.39\linewidth}
\vspace{0pt}\small\textbf{Evidence detail}\par\smallskip
\includegraphics[width=0.56\linewidth,trim={925.0000bp 880.0000bp 496.0000bp 990.0000bp},clip]{images/reviewed_case_vstar_29.jpg}
\end{minipage}
\par\medskip
\begin{minipage}{\linewidth}
\textbf{Question.} What is the color of the man's cap?\par\smallskip
A. white\qquad B. red\qquad C. gray\qquad D. blue\par\medskip
\textbf{Reference answer.} B (red)\par\medskip
\setlength{\tabcolsep}{8pt}\renewcommand{\arraystretch}{1.4}
\begin{tabular}{>{\raggedright\arraybackslash}p{0.18\linewidth}>{\raggedright\arraybackslash}p{0.57\linewidth}>{\raggedright\arraybackslash}p{\dimexpr0.25\linewidth-6\tabcolsep\relax}}
\toprule Method & Recorded answer text & Outcome \\
\midrule
\rowcolor{red!7}HiPrune & \texttt{(A)} & Incorrect \\
\rowcolor{red!7}EPIC & \texttt{A} & Incorrect \\
\rowcolor{green!7}LT-OPD & \texttt{B} & Correct \\
\bottomrule
\end{tabular}
\end{minipage}
\caption{V$^*$Bench item 29. A red cap occupies a small part of the poolside scene. Both baselines select white; LT-OPD correctly selects red.}

\label{fig:case_cap}
\end{figure}

\clearpage
\subsection{Distinguishing a Document Date from a Mentioned Event Date}
\begin{figure}[H]
\centering
\begin{minipage}[t]{0.53\linewidth}
\vspace{0pt}\includegraphics[width=\linewidth]{images/reviewed_case_ocrbench_512.jpg}
\end{minipage}\hfill
\begin{minipage}[t]{0.43\linewidth}
\raggedright\vspace{0pt}\small\textbf{Document-date detail}\par\smallskip
\includegraphics[width=\linewidth,trim={743.0000bp 1947.0000bp 708.0000bp 266.0000bp},clip]{images/reviewed_case_ocrbench_512.jpg}
\end{minipage}
\par\medskip
\begin{minipage}{\linewidth}
\textbf{Question.} When is the document dated?\par\medskip
\textbf{Reference answer.} January 9, 1961\par\medskip
\setlength{\tabcolsep}{8pt}\renewcommand{\arraystretch}{1.4}
\begin{tabular}{>{\raggedright\arraybackslash}p{0.18\linewidth}>{\raggedright\arraybackslash}p{0.57\linewidth}>{\raggedright\arraybackslash}p{\dimexpr0.25\linewidth-6\tabcolsep\relax}}
\toprule Method & Recorded answer text & Outcome \\
\midrule
\rowcolor{red!7}HiPrune & \texttt{The document is dated **March 13, 1975**.} & Incorrect \\
\rowcolor{red!7}EPIC & \texttt{The document is dated December 1, 1960.} & Incorrect \\
\rowcolor{green!7}LT-OPD & \texttt{January 9, 1961} & Correct \\
\bottomrule
\end{tabular}
\end{minipage}
\caption{OCRBench item 512 (DocVQA). LT-OPD correctly returns the letter's header date, January 9, 1961. EPIC returns the meeting date mentioned in the body, December 1, 1960; HiPrune returns a different date.}
\label{fig:case_date}
\end{figure}

\clearpage

\end{document}